\documentclass[journal]{IEEEtran}
\usepackage{amsmath,amsfonts}
\usepackage{algorithmic}
\usepackage{array}
\usepackage[caption=false,font=normalsize,labelfont=sf,textfont=sf]{subfig}
\usepackage{textcomp}
\usepackage{stfloats}
\usepackage{url}
\usepackage{verbatim}
\usepackage{graphicx}

\newcommand{\blockcomment}[1]{}
\usepackage{multirow}
\usepackage{color}
\usepackage{colortbl}
\usepackage{xcolor, soul}
\newcommand\SLASH{\char`\\}
\newcommand{\overbar}[1]{\mkern 1.5mu\overline{\mkern-1.5mu#1\mkern-1.5mu}\mkern 1.5mu}
\graphicspath{{Images/}} 

\newcolumntype{P}[1]{>{\centering\arraybackslash}p{#1}}
\newcolumntype{M}[1]{>{\centering\arraybackslash}m{#1}}
\definecolor{Gray}{gray}{0.85}
\definecolor{LightGray}{gray}{0.95}

\newtheorem{definition}{Definition}[section]
\makeatletter
\renewcommand*{\@opargbegintheorem}[3]{\trivlist
      \item[\hskip \labelsep{\bfseries #1\ #2}] \textbf{(#3)}\ \itshape}
\makeatother

\newcommand{\PU}{\mu^{{U}}}
\newcommand{\PURAW}{\tilde{\mu}^{{U}}}
\newcommand{\CC}{\mu^{C}}
\newcommand{\CCRAW}{\tilde{\mu}^{{C}}}

\newcommand{\AB}{}
\renewcommand{\textit}{}
\newcommand{\mathcolorbox}[2]{\colorbox{#1}{$\displaystyle #2$}}
\usepackage{hyperref}
\usepackage[capitalise]{cleveref}
\title{Class Uncertainty: \\ A Measure to Mitigate Class Imbalance}
\begin{document}
\author{Zeynep Sonat Baltaci$^{1,2}$, Kemal Oksuz$^{3}$, Selim Kuzucu$^{1}$, Kivanc Tezoren$^{1}$, Berkin Kerim Konar$^{1}$, \\Alpay Ozkan$^{1}$, Emre Akbas$^{1,4,\dagger}$, Sinan Kalkan$^{1,4,\dagger}$\\
  $^1$Dept. of Computer Engineering, METU, Ankara, Turkey \\
  $^2$LIGM, Ecole des Ponts, Univ Gustave Eiffel, CNRS, Marne-la-Vallée, France \\
  $^3$Five AI Ltd., United Kingdom \\
  $^4$Center for Robotics and Artificial Intelligence (ROMER), METU, Ankara, Turkey \\
  sonat.baltaci@enpc.fr,  kemal.oksuz@five.ai, \\
  \{selim.kuzucu, kivanc.tezoren, berkin.konar, alpay.ozkan, eakbas, skalkan\}@metu.edu.tr
\thanks{Manuscript received October 3, 2024.}}
\maketitle

\markboth{Under Review}%
{Class Uncertainty: A Measure to Mitigate Class Imbalance}
\def\thefootnote{$\dagger$}\footnotetext{Equal contribution for senior authorship.}\def\thefootnote{\arabic{footnote}}
\begin{abstract}
Class-wise characteristics of training examples affect the performance of deep classifiers. A well-studied example is when the number of training examples of classes follows a long-tailed distribution, a situation that is likely to yield sub-optimal performance for under-represented classes. This class imbalance problem is conventionally addressed by approaches relying on the class-wise cardinality of training examples, such as data resampling. In this paper, we demonstrate that considering solely the cardinality of classes does not cover all issues causing class imbalance. To measure class imbalance,  we propose \textsc{Class Uncertainty} as the average predictive uncertainty of the training examples,  and we show that this novel measure captures the differences across classes better than cardinality. We also curate SVCI-20 as a novel dataset in which the classes have equal number of training examples but they differ in terms of their hardness; thereby causing a type of class imbalance which cannot be addressed by the approaches relying on cardinality. We incorporate our \textsc{Class Uncertainty} measure into a diverse set of ten class imbalance mitigation methods to demonstrate its effectiveness on long-tailed datasets as well as on our SVCI-20. Code and datasets will be made available.
\end{abstract}

\begin{IEEEkeywords}
long-tailed visual recognition, class imbalance, class uncertainty, predictive uncertainty
\end{IEEEkeywords}

\section{Introduction}
\label{section:intro}
The performance of data-driven, learning-based methods relies heavily on dataset characteristics. One characteristic that has received significant attention is the unequal cardinality of examples per class\footnote{Throughout the paper, we use  \textit{cardinality} to refer to the number of training examples for a class.} as in~\cite{Peng_2020_CVPR,kang2019decoupling, cui2019classbalancedloss, han2018coteaching, Park_2021_ICCV, Fernando2021-tl,ldam,khan2019striking,menon2021longtail,Samuel_2021_ICCV}. This attention originates from the fact that the frequencies of object classes follow a long-tailed distribution in nature, which also manifests itself in many visual recognition datasets~\cite{openlongtailrecognition, yang2022survey, oksuz2020imbalance, Krawczyk2016LearningFI, Swagatam2022, cheong2021hitchhiker}. 
Accordingly, the long-tailed distribution of cardinalities is generally considered as the major reason for many performance issues encountered on such datasets. 
This common phenomenon can be more accurately described as the ``cardinality imbalance”. 
 \AB{The issue of ``imbalance" arises when the training set is imbalanced while the test set is balanced. This is a common practical challenge, since in real-world scenarios, during inference, we want our models to perform well for all classes, and not just for some of the classes. Academic datasets (such as CIFAR-LT~\cite{cui2019classbalancedloss}) address this by creating a balanced test set, despite the imbalances in the training set. However, as we emphasize in this paper, cardinality is just one facet of imbalance, and a balanced test set may not ensure equal difficulty (hardness) levels among classes.} 
\begin{figure*}[hbt!]
        \centering
        \subfloat[Gains of our \textsc{Class Uncertainty}]{\includegraphics[width=0.3\textwidth]{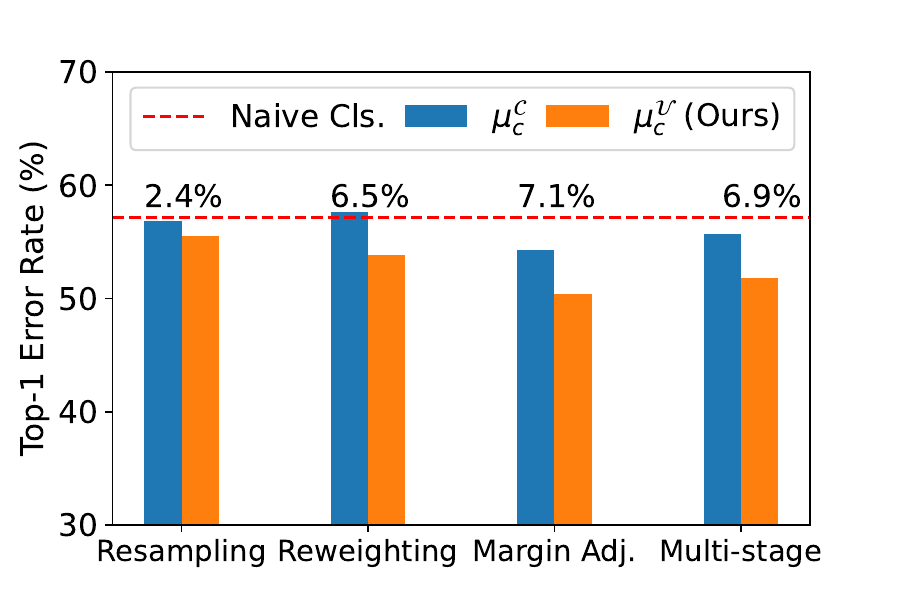}}\hspace*{1cm}
        \subfloat[Na\"ive classifier vs. different ways of handling imbalance]{\includegraphics[width=0.6\textwidth]{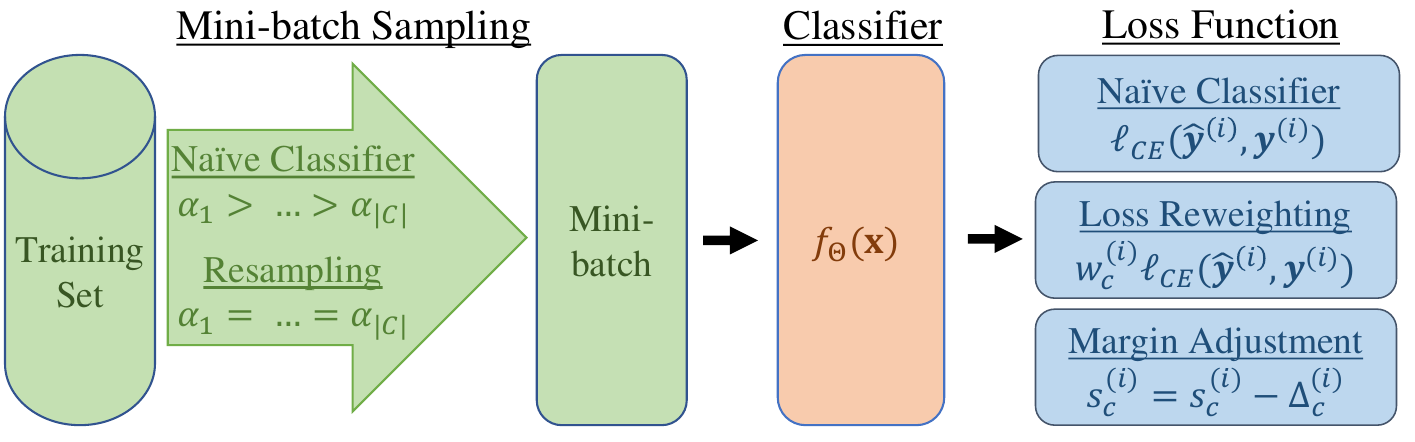}}
        \caption{
        \textbf{(a)} Using our \textsc{Class Uncertainty} improves top-1 error rate of all aforementioned methods (CIFAR-100-LT with an IR of 50 using ResNet-32). In particular, we obtain (i) the sampling probability of each class in resampling methods; (ii) the weights of the loss for each class; (iii) the margins to be enforced around each class; or (iv) again sampling probability in the second-stage of multi-stage training strategy using \textsc{Class Uncertainty}. The numbers on top of the histograms show  relative gain over cardinality-based methods. Baselines: Progressively-balanced sampling~\cite{kang2019decoupling}, class-balanced Focal Loss~\cite{cui2019classbalancedloss}, LDAM with reweighting~\cite{ldam} and deferred resampling.
        \textbf{(b)} Na\"ive classifier is not robust to class imbalance, giving rise to a plethora of mitigation methods: (1) ``resampling" samples a balanced set of examples; (2) ``reweighting'' assigns a weight for each class ($w_c^{(i)}$); (3) ``margin-adjustment'' methods assign different margins ($\Delta_c^{(i)}$) to the logits ($s_c^{(i)}$) of different classes; and (4) ``multi-stage training'' first trains a na\"ive classifiers  followed by a resampling or reweighting method to re-train or fine-tune the classifier. These methods generally rely on the cardinality of the classes. 
        }
        \label{fig:teaser}
\end{figure*}
Although cardinality imbalance is the most evident type of imbalance to observe, it does not completely capture or explain the spectrum of factors that cause performance gaps among classes.
Other contributing factors include the hardness of the examples or classes~\cite{focal,OHEM} and the discrepancy between the training objective and the evaluation measure~\cite{APLoss,aLRPLoss}. If such factors, which we will refer to generally  as ``class imbalance’’, are not addressed properly, they lead to sub-optimal performance for certain classes.

In this paper, we challenge the aforementioned convention to rely on cardinality to alleviate the class imbalance and make the following main contributions:
\begin{itemize}
    \item To set an important milestone in the class imbalance literature, we define what a class imbalance measure is and outline the important features it should have. 
    \item We introduce a new measure, \textsc{Class Uncertainty}, as the average predictive uncertainty of training examples in a class. We extensively show its effectiveness compared to cardinality. Specifically, incorporating our measure into a diverse set of ten methods from resampling, loss reweighting, margin adjustment, and multi-stage training; we observe better or on-par performance compared to class cardinality on long-tailed datasets as highlighted in~\cref{fig:teaser}(a). As an example, \textsc{Class Uncertainty} drastically reduces the top-1 error of the recent LDAM method by $3.5-4.0 \%$ on the CIFAR-100-LT dataset.
    \item To facilitate studies with a broader perspective than cardinality, we curate a new dataset, SVCI-20, where classes have equal cardinality but different hardness.
\end{itemize}

We present these contributions in the following sections:~\cref{sec:background} includes the background and discusses why minimizing the negative log-likelihood is not robust to class imbalance. It also provides a simple taxonomy of existing imbalance mitigation methods with a brief discussion of such methods. \cref{sec:analysis} defines what a class imbalance measure is, analyzes using class cardinality as an imbalance measure, and proposes our \textsc{Class Uncertainty}. \cref{sec:mitigate} demonstrates how to incorporate our measure into existing mitigation methods. \cref{section:exps} presents our experimental results for class uncertainty on long-tailed datasets and our novel semantically-imbalanced dataset, SVCI-20. Finally,~\cref{sec:discuss} concludes the paper.
\section{Background}
\label{sec:background}
\subsection{Is Na\"ive Classifier Robust to Class Imbalance?}
\label{subsec:classification}
Consider a classifier that minimizes the negative log-likelihood (NLL) of training examples, which is the prevalent approach in the deep learning literature~\cite{VGG,GoogLeNet,inceptionv2,ResNet,ResNext,DenseNet}.   We term this classifier as the \textit{na\"ive classifier}. Such a classifier, $f_{\Theta}:\mathcal{X}\rightarrow\mathcal{Y}$, maps an image $\mathbf{x}^{(i)} \in \mathcal{X}$ to a categorical distribution $\mathbf{\hat{y}}^{(i)} \in \mathcal{Y}$ such that $\hat{y}^{(i)}_c$, $c$-th element of $\mathbf{\hat{y}}^{(i)}$, represents the predicted probability for class $c\in \mathcal{C}$. Typically, $\hat{y}^{(i)}_c$ is obtained through the softmax function: 
\begin{equation}
\label{eq:softmax}
\hat{y}^{(i)}_c = \frac{\exp({s^{(i)}_c})}{\sum_{c'\in \mathcal{C}} \exp({s^{(i)}_{c'}})},
\end{equation}
with $s^{(i)}_c$ being the predicted logit for class $c$. Given $\mathbf{y}^{(i)} \in \{0,1\}^{|\mathcal{C}|} \subseteq \mathcal{Y}$, the label of the $i$th example, $f_{\Theta}(\cdot)$ is conventionally trained by minimizing Cross Entropy Loss ($\ell_{CE}(\cdot, \cdot)$) using mini-batch Stochastic Gradient Descent (SGD)~\cite{PMLR}. Formally, the loss of a mini-batch is:
\begin{equation}
\label{eq:loss}
\mathcal{L} = \frac{1}{m} \left( \sum_{i=1}^m \ell_{CE}(\mathbf{y}^{(i)}, \mathbf{\hat{y}}^{(i)}) \right),  
\end{equation}
where $m$ is the number of examples in the mini-batch.
By definition, minimizing NLL over training using mini-batch SGD data does not prioritize any example; thereby making the na\"ive classifier sensitive to the class characteristics. As a result, the larger the cardinality a class has, the more it is represented in the computed NLL. Specifically, as illustrated in~\cref{fig:teaser}(b), the random sampling of mini batches would naturally cause the sampling probability of an example from class $c$ ($\alpha_c$) to be proportional to the cardinality of the class. To state more formally,  $\alpha_1>\alpha_2>...> \alpha_{|\mathcal{C}|}$ holds true, assuming that the class indices are ordered with respect to (wrt.) the cardinality in decreasing order. As a result, the loss function does not focus on the examples that are under-represented either. Consequently, in the case of an imbalance between different classes, training is prone to overfit to the over-represented classes. 

A prominent failure example of the na\"ive classifier is observed in training object detectors~\cite{oksuz2020imbalance}. While training typical object detectors, there are prohibitively more negative examples from the background compared to very few positive examples from the foreground (i.e., objects). Consequently, unless an imbalance mitigation method is employed while training an object detector, the na\"ive classifier completely ignores the positive examples, the under-represented classes in this context~\cite{focal,aLRPLoss}. Therefore, the na\"ive classifier is not robust against class imbalance, a problem that remains open despite a plethora of mitigation techniques.

\subsection{Mitigating Class Imbalance}
\label{subsec:relatedwork}
We categorize the existing class imbalance mitigation techniques into four groups. An overview is given in~\cref{fig:teaser}(b).
\subsubsection{Resampling}
This set of methods addresses class imbalance during mini-batch sampling. One standard approach is to equalize the sampling probabilities of classes:
\begin{equation}
    \label{eq:resampling}
    \alpha_1 = \alpha_2 = ... = \alpha_{|\mathcal{C}|},
\end{equation}
known as classical oversampling or class-balanced (CB) resampling~\cite{Shen2016RelayBF}. As this approach is prone to overfitting in favor of tail classes, several mitigation strategies have been proposed. Peng et al.~\cite{Peng_2020_CVPR} combined CB and random sampling with a weight parameter to obtain $\alpha_c$ whereas Kang et al.~\cite{kang2019decoupling} introduced progressive-balanced (PB) resampling, which initializes $\alpha_c$s with random sampling before gradually being replaced by CB resampling. 
\subsubsection{Loss Reweighting}
This common group of approaches~\cite{csce, focal, cui2019classbalancedloss, robust, han2018coteaching,  JamalLongtail_DA, Park_2021_ICCV, Fernando2021-tl} reweights the loss values of the examples to promote certain classes or examples:
\begin{equation}
\label{eq:lossweighting}
\mathcal{L} = \frac{1}{m} \sum_{i=1}^m w^{(i)}_c \ell_{CE}(\mathbf{\hat{y}}^{(i)}, \mathbf{y}^{(i)}),
\end{equation}
where $w^{(i)}_c$ is the weight of the example $i$ from class $c$. Examples include the Cost-sensitive Cross Entropy (CSCE) Loss~\cite{csce} relying on the inverse of the class cardinalities as $w^{(i)}_c$; and  the Class-balanced Loss~\cite{cui2019classbalancedloss} introducing an ``effective'' number of examples for each class, which practically smooths the weights of CSCE. An alternative approach, Focal Loss~\cite{focal}, assigns an example-specific weight based on the prediction confidence of the classifier as an indicator of hardness.
\subsubsection{Margin Adjustment}
These relatively recent methods~\cite{ldam,khan2019striking,menon2021longtail,Kini2021LabelImbalancedAG,Samuel_2021_ICCV} introduce margin terms to the logit of each class, effectively demotizing over-represented classes. This is achieved by subtracting a class-specific margin $\Delta^{(i)}_c$ from the predicted logit $s^{(i)}_c$ while computing~\cref{eq:softmax}. Specifically, we generalize the existing margin adjustment methods as:
\begin{equation}
\label{eq:softmax_margin}
\hat{y}^{(i)}_c = \frac{\exp({s^{(i)}_c - \Delta^{(i)}_c})}{\exp({s^{(i)}_c - \Delta^{(i)}_c}) + \sum_{k\in \mathcal{C}\SLASH c} \exp({s^{(i)}_k- \Delta^{(i)}_k \beta})},
\end{equation}
where $\beta \in \{0,1\} $ determines whether or not to take into account the margins assigned to other classes for class $c$. The examples include: (i) LDAM~\cite{ldam} with $\beta=0$, $\Delta^{(i)}_c = \tau /N_c^{1/4}$ and $\tau$ is a hyper-parameter; (ii) Logit-adjusted Loss~\cite{menon2021longtail} with $\beta = 1$ and $\Delta^{(i)}_c = - \kappa \log (N_i / \sum_{k=1}^\mathcal{C} N_j)$ and $\kappa$ being its hyper-parameter; and (iii) Distributional Robustness Loss (DRO)~\cite{Samuel_2021_ICCV}, which, unlike other methods, enforces margins at the feature-level. 

\subsubsection{Multi-stage Training}
Multi-stage training methods are based on the idea of decoupling learning between representation and classification~\cite{kang2019decoupling}. Specifically, such methods first train the classifier without a mitigation method and subsequently continue by training only the last layer~\cite{kang2019decoupling} or the whole network~\cite{ldam, zhang2020tricks} while leveraging an imbalance mitigation method such as CB resampling or CSCE. Several methods are built on this training strategy due to its simplicity and effectiveness~\cite{DistributionAlignment,Samuel_2021_ICCV,menon2021longtail,ImprovingCalibration,Park_2021_ICCV}. 

\textbf{Discussion} As evidenced by the overview of the related work above, \emph{existing methods mostly rely on ``class cardinality'' to measure class imbalance in a dataset.} Specifically, among the aforementioned methods, (i) resampling methods; (ii) loss reweighting methods except Focal Loss; (iii) margin adjustment methods except a specific setting of DRO-LT Loss; and (iv) multi-stage training methods are all based on class cardinality.
\emph{In this paper, we challenge this ubiquitously adopted approach of using ``class cardinality'' as the sole basis for imbalance mitigation.}
\section{\textsc{Class Uncertainty}: A Novel Measure of Class Imbalance}
\label{sec:analysis}
In this section, following our viewpoint of a class imbalance measure, we present an analysis on class cardinality, remark its limitations and propose \textsc{Class Uncertainty} to measure and mitigate class imbalance.
\subsection{Measuring Class Imbalance}
\label{subsec:perspective}
As summarized in the previous section, the current literature on class imbalance is heavily based on class cardinality, making it the defacto measure of class imbalance. A better measure promises benefits for different types of mitigation techniques. 
\begin{definition}[\textsc{Class Imbalance Measure}]
{A class imbalance measure is a function $\mathrm{\mu}$ that takes in a dataset $\mathcal{D} \subseteq \mathcal{X} \times \mathcal{Y}$ with $|\mathcal{C}|$ classes as input and returns a $|\mathcal{C}|$-dimensional vector. For convenience, we reuse $\mu$ to denote the output of this function, which is in $\mathbb{R}^{|\mathcal{C}|}$.}
\end{definition}
Moreover, we introduce the following two important features (IF)  that a class imbalance measure should have:
\begin{itemize}
    \item \textbf{(IF1) Ability to capture different amounts of imbalance.} An imbalance measure should be able to quantify how well a class is represented by its examples. As such, we expect a measure to:
    \begin{itemize}
        \item (\textbf{IF1a}) be higher for an under-represented class than an over-represented class, and
        \item (\textbf{IF1b}) decrease for the over-represented classes (or increase for the under-represented classes) as the dataset becomes more imbalanced. 
    \end{itemize}    
    \item \textbf{(IF2) Robustness to the amount of ineffective examples.} An imbalance measure should be robust to the presence of `ineffective' examples\footnote{Following Cui et al.~\cite{cui2019classbalancedloss}, an example can be named effective if it covers sufficiently different space than other examples.} that might have populated the dataset. In essence, we desire our measure to capture the imbalance not in terms of quantity (cardinality), but quality.
\end{itemize}
These important features allow us to gain more insight into class cardinality and our proposed measure. In these analyses, for (\textbf{IF1a}) it is not trivial to identify over- and under-represented classes, as we argue that class imbalance is not limited to cardinality imbalance. Therefore, we use the class-wise test set errors of the na\"ive classifier as a valid proxy to the underlying imbalance. For (\textbf{IF1b}), we change the imbalance ratio (IR) of the dataset and plot how the different measures vary. As for (\textbf{IF2}), we duplicate the training examples in favor of under-represented classes and exploit these duplicates as ineffective examples. Please refer to~\cref{subsec:imp_details} for details.

\begin{figure*}[hbt!]
        \centering
        \subfloat[$\CC$ vs. $\PU$ on CIFAR-10-LT]{\includegraphics[width=0.3\textwidth]{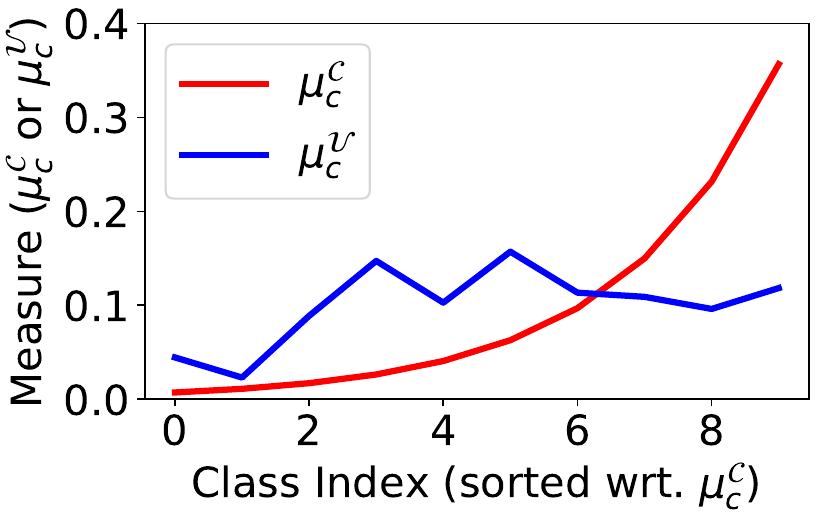}}
       \subfloat[$\CC$ vs. $\PU$ on CIFAR-100-LT]{\includegraphics[width=0.3\textwidth]{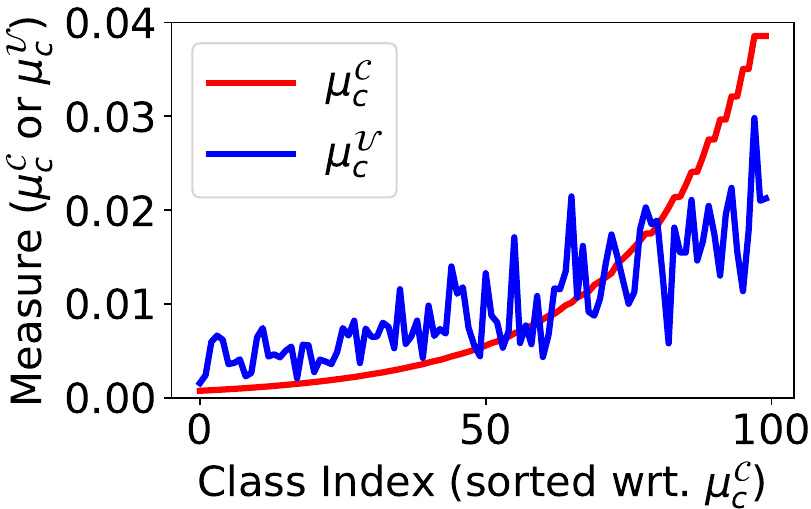}}
       \hfill
        \subfloat[\textbf{(IF1a)} for $\CC_c$]{\includegraphics[width=0.3\textwidth]{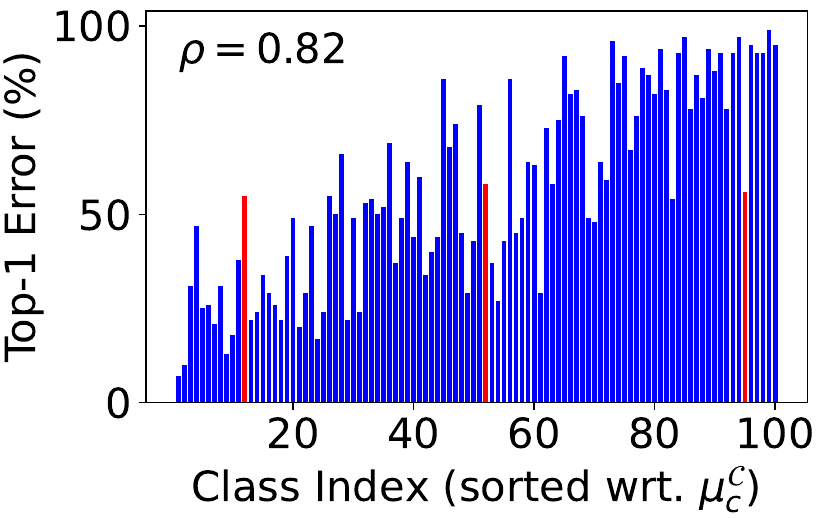}}
        \subfloat[\textbf{(IF1b)} for $\CC_c$]{\includegraphics[width=0.3\textwidth]{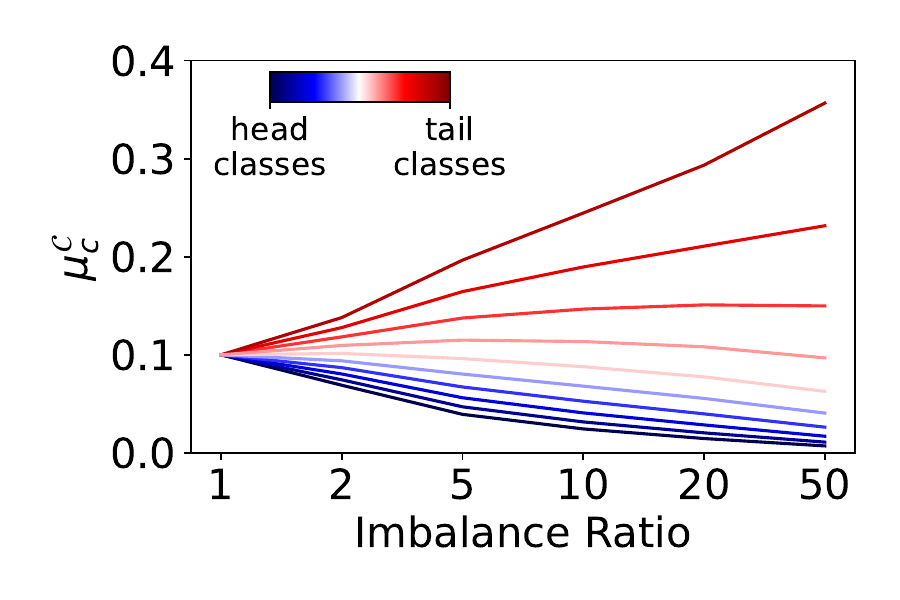}}
        \subfloat[\textbf{(IF2)} for $\CC_c$]{\includegraphics[width=0.3\textwidth]{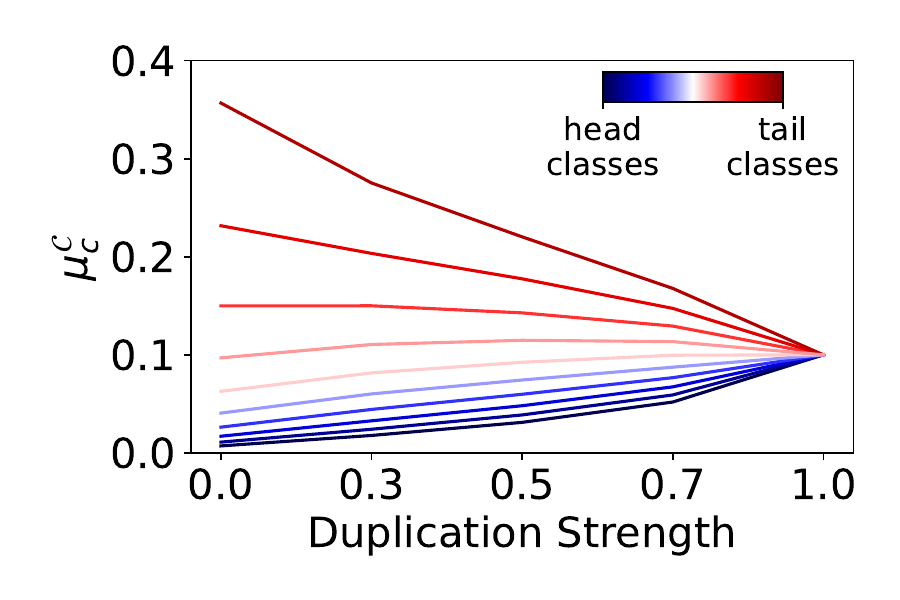}}
        \hfill
        \subfloat[\textbf{(IF1a)} for $\PU_c$]{\includegraphics[width=0.3\textwidth]{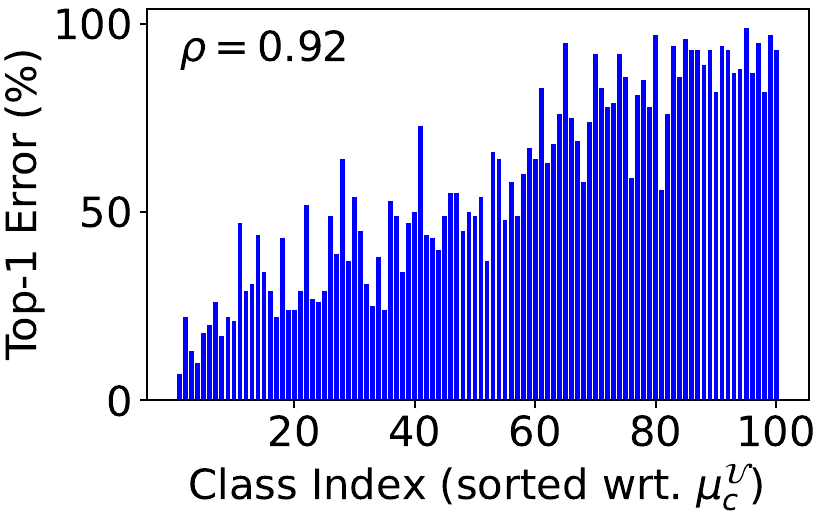}}
        \subfloat[\textbf{(IF1b)} for $\PU_c$]{\includegraphics[width=0.3\textwidth]{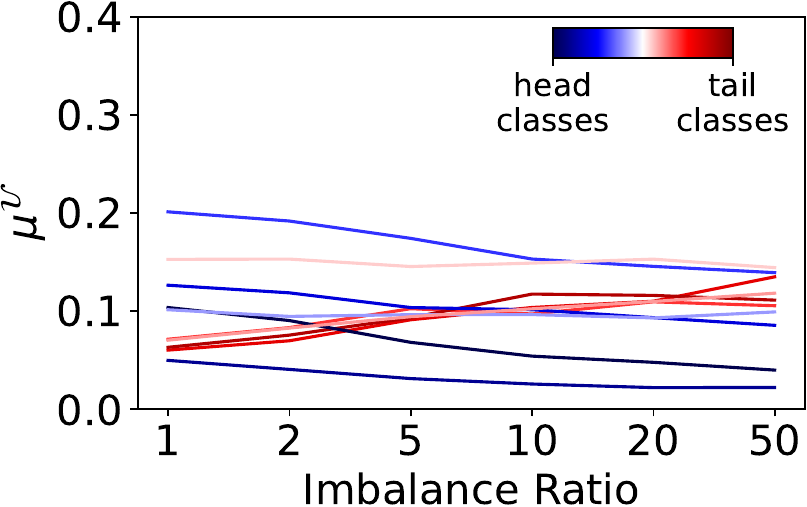}}
        \subfloat[\textbf{(IF2)} for $\PU_c$]{\includegraphics[width=0.3\textwidth]{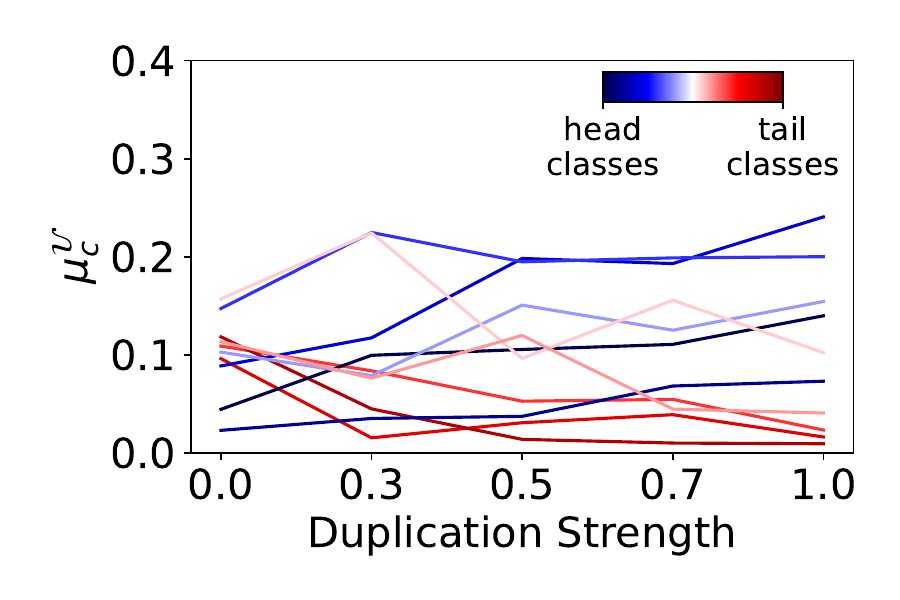}}
        \caption{ \textbf{(a,b)} \textsc{Class Cardinality Imbalance Measure} ($\CC_c$) vs. \textsc{Class Uncertainty} ($\PU_c$) on CIFAR-10-LT and CIFAR-100-LT with an IR of 50. While $\CC_c$ is entirely determined by cardinality (from left to right, class cardinality decreases), $\PU_c$ presents more diverse values. \textbf{(c-h)} An analysis of $\CC_c$ and $\PU_c$ in terms of important features. For capturing the amount of imbalance among classes ((c) and (f)), Spearman correlation coefficient ($\rho$) between the imbalance measure and the test accuracy of a na\"ive classifier on CIFAR-100-LT is notably higher for $\PU_c$ compared to $\CC_c$, which implies that $\PU_c$ captures imbalance better. As IR increases, both can capture the change in IR in (d) and (g) by promoting the tail classes (in red) or demoting the head ones (in blue), while the relative change across classes in $\CC_c$ is more drastic. In (e) and (h), $\CC_c$ is highly sensitive to the duplicated training examples, which do not contain new information. $\PU_c$ is relatively less sensitive especially after the duplication strength of $0.3$. Therefore, $\PU_c$ is a better alternative to $\CC_c$ when considering these important features.
        }
        \label{fig:analysis}
\end{figure*}

\subsection{\textsc{Class Cardinality Imbalance Measure} and Its Analysis}
\label{subsec:cardinality}
Considering that cardinality-based methods~\cite{Shen2016RelayBF,csce,cui2019classbalancedloss} promote classes with fewer examples during training, we formulate the underlying measure as follows:
\begin{definition}[\textsc{Class Cardinality Imbalance Measure}]
\textsc{Class Cardinality Imbalance Measure}, $\CC$, is an imbalance measure that is inversely proportional to $N_c$, the number of training examples in class $c$. Thus, the imbalance measured for the class $c$ is:
\begin{equation}
\label{eq:cardinality_}
\mathcolorbox{white}{
    \CCRAW_c = \frac{1}{N_c},
}
\end{equation}
\AB{which is then normalized as:}
\begin{equation}
\label{eq:cardinality}
\mathcolorbox{white}{
\CC_c =  \frac{{\CCRAW_{c}}}{\sum_{c' \in C}{\CCRAW_{c'}}}.    
}
\end{equation}
\AB{such that $\sum_{c \in \mathcal{C}} \CC_c = 1$.} 
\end{definition}
Normalization enables us to provide an analysis independent of the range of the measure. Hence, if the class $c$ has fewer examples, then the corresponding $\CC_c$ increases relative to the others. As an example, the red curves in~\cref{fig:analysis}(a,b) demonstrate $\CC_c$ for CIFAR-10-LT and CIFAR-100-LT~\cite{cui2019classbalancedloss}.

\textbf{Analysis} Here we analyze \textsc{Class Cardinality Imbalance Measure} based on the important features from~\cref{subsec:perspective}. In order to examine \textbf{(IF1a)}, the extent to which $\CC_c$ captures the relations between different classes, we plot class indices wrt. $\CC_c$ vs. the test set error of class $c$ in~\cref{fig:analysis}(c). The figure demonstrates that the error of a class \textit{generally increases} as $\CC_c$ increases. However, there are some classes with very different number of examples but similar test errors. For example, three classes represented by the red histograms in~\cref{fig:analysis}(c) have very similar errors, but they have 323, 66 and 12 examples in the training set respectively; implying that $\CC_c$ does not necessarily quantify the imbalance in a class perfectly. In terms of \textbf{(IF1b)},~\cref{fig:analysis}(d) shows how $\CC_c$ changes as the IR increases; suggesting that the relative change between the under- (red tones) and over-represented classes (blue tones) is drastic as the IR increases. Finally, for \textbf{(IF2)},~\cref{fig:analysis}(e) suggests that $\CC_c$ is very sensitive to duplicates and ineffective samples. Overall, while $\CC_c$ has benefits and has been used extensively in the literature, our brief analysis shows that it has limitations and offers room for improvement.

\subsection{Our Proposal: \textsc{Class Uncertainty} to Measure \& Mitigate Imbalance}
Considering the limitations of \textsc{Class Cardinality Imbalance Measure} (\cref{subsec:cardinality}), we seek to design a measure of class imbalance that not only considers the cardinality of a class but also reflects semantic information such as hardness, etc. This motivation leads us to using predictive uncertainty as a promising measure considering that it captures both epistemic uncertainty reflecting the lack of data at a point in the input space and aleatoric uncertainty, representing ambiguity and noise. More specifically, we argue that predictive uncertainty is suitable as a measure of class imbalance for the following two main reasons:
\begin{itemize}
    \item Defined as a measure of lack of data, the epistemic component of predictive uncertainty, by definition, is a suitable measure to capture the imbalance in data across classes. An important benefit of epistemic uncertainty is that the notion of sufficiency of data is not determined by the number of examples, but rather with the behavior of the machine learning model.
    \item Determining the noise or the ambiguity, the aleatoric component of the predictive uncertainty provides information of the example itself. To illustrate, Kendall and Gal~\cite{whatuncertainties} use the depth regression task, a closely related computer vision task, to elaborate on aleatoric uncertainty. They demonstrate that this type of uncertainty captures inherently difficult aspects, such as large depths and reflective surfaces. Hence, this and similar observations on the aleatoric uncertainty align with our expectation to include the semantic hardness of example in the imbalance measure. \AB{Our approach aligns with that of reweighting the loss with $(1-\hat{p}^{(i)})$ in Focal Loss~\cite{focal}, a prominent method of class imbalance, where $(1-\hat{p}^{(i)})$ is in fact a measure of uncertainty~\cite{Gal2016UncertaintyID}.}
\end{itemize}
Since deep learning-based approaches are over-confident in their predictions~\cite{FocalLoss_Calibration}, we train $T$ different models on the same training set, where typically $T=5$, and quantify the uncertainty given the $T$ predictions of these models, known as Deep Ensembles (DE)~\cite{de}\AB{\footnote{\AB{See Supp. Mat. for an analysis and discussion on uncertainty quantification.}}}. Formally, denoting the prediction vector of the $t$th model by $\mathbf{\hat{p}}^{(i),t}$ and the probability for class $c$ by $\hat{p}_c^{(i),t} \in \mathbf{\hat{p}}^{(i),t}$, the predictive uncertainty of the deep ensemble in example $i$, ${u}^{(i)}$, can be quantified as,
\begin{equation}
    {u}^{(i)} = - \sum_{c \in \mathcal{C}} \bar{p}_{c}^{(i)} \log \bar{p}_{c}^{(i)},
    \label{eq:uncertainty}
\end{equation}
where $\bar{p}_c^{(i)} = \frac{1}{T} \mathcolorbox{white}{\sum_{t \in T}} \hat{p}_c^{(i),t}$.

To adopt predictive uncertainty as a class imbalance measure and obtain a \textit{class-level} imbalance measure given \textit{example-level} predictive uncertainties, we aggregate example-level predictive uncertainties from the training set simply by the class-wise average:
\begin{definition}[\textsc{Class Uncertainty Imbalance Measure}]
{\textsc{Class Uncertainty Imbalance Measure}, $\PU$, is the average predictive uncertainties of the training examples for each class. Note that unlike $\CC$, $\PU$ is a function of not only the dataset $\mathcal{D}$ but also a model. Specifically, given $u^{(i)}$ as the predictive uncertainty (defined in~\cref{eq:uncertainty})  of the $i^{th}$ training example of class $c$ and $N_c$ as the cardinality of class $c$, the unnormalized imbalance in class $c$ is:}
\begin{equation}
    {\PURAW_{c}} =  \frac{1}{N_c}\sum_{i=1}^{N_c} {u^{(i)}},
    \label{eq:cw_uncertainty_}
\end{equation}
which is then normalized similar to~\cref{eq:cardinality}:
\begin{equation}
    \PU_c =  \frac{{\PURAW_{c}}}{\sum_{c' \in C}{\PURAW_{c'}}}.
    \label{eq:cw_uncertainty}
\end{equation}
\end{definition}
Blue curves in~\cref{fig:analysis}(a,b) illustrate $\PU_c$ for CIFAR-10-LT and CIFAR-100-LT in which $\PU_c$ presents more diverse values than $\CC_c$ that are not entirely determined by the cardinality of the classes.

\textbf{Analysis} We now investigate how \textsc{Class Uncertainty} behaves wrt. our important features compared to \textsc{Class Cardinality Imbalance Measure}. While quantifying uncertainty, we use a Deep Ensemble (DE)~\cite{de} whose effectiveness in yielding reliable and calibrated uncertainty estimates is proven over a wide range of datasets and models~\cite{de,regmixup,regressionunc}. Firstly for \textbf{(IF1a)},~\cref{fig:analysis}(f) depicts how class-wise test error changes wrt.  \textsc{Class Uncertainty}. In a quantitative summary, the Spearman correlation coefficient $\rho$ between class-wise test errors and the \textsc{Class Uncertainty} measure is $0.92$, while it is $0.82$ for \textsc{Class Cardinality Imbalance Measure} in~\cref{fig:analysis}(c). This suggests that \textsc{Class Uncertainty} captures the imbalance between classes better than \textsc{Class Cardinality Imbalance Measure}. Secondly,~\cref{fig:analysis}(g) demonstrates for \textbf{(IF1b)} that as the IR increases, the imbalance assigned to under-represented classes increases while that of over-represented ones decreases, as expected from a good imbalance measure. Finally, to compare the robustness of the measures for \textbf{(IF2)}, comparing~\cref{fig:analysis}(e) and~\cref{fig:analysis}(h) suggests that \textsc{Class Uncertainty} is affected from duplicate examples less than \textsc{Class Cardinality Imbalance Measure} and is a more robust measure. This analysis suggests that \textsc{Class Uncertainty} is a better alternative to \textsc{Class Cardinality Imbalance Measure} to quantify the class imbalance.
\section{Mitigating Class Imbalance Using \textsc{Class Uncertainty}}
\label{sec:mitigate}
\begin{figure*}
    \centering
    \includegraphics[width=\textwidth]{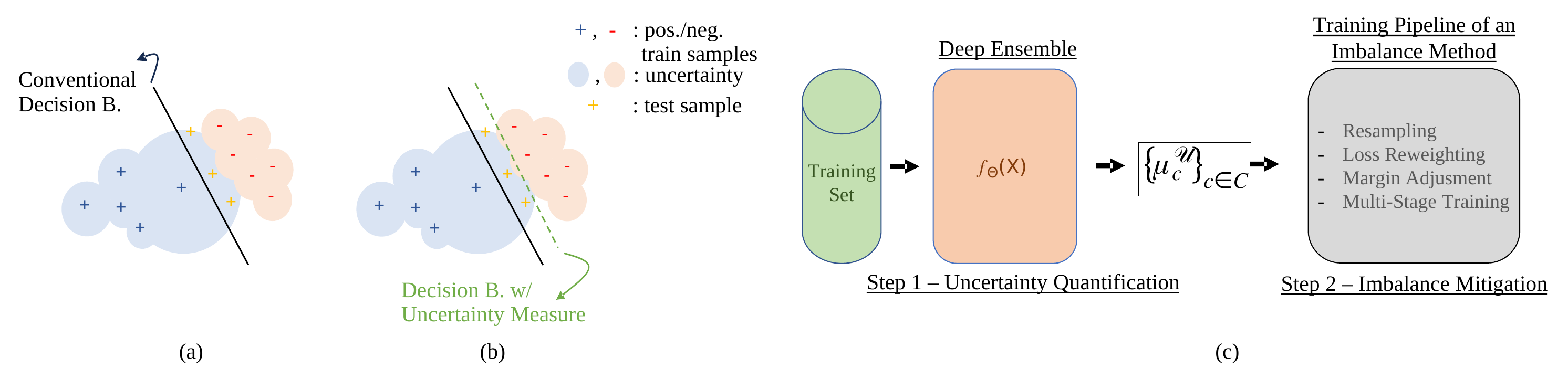}
    \caption{\AB{\textbf{(a)} Conventional imbalance mitigation methods ignore variance of samples while estimating the decision boundary. \textbf{(b)} Our motivation is to promote the learning of classes with higher uncertainty, effectively pushing the decision boundary away from such classes. 
    \textbf{(c)}} High-level overview of our strategy. We first obtain \textsc{Class Uncertainty} ($\PU_c$) as the measure of imbalance using the predictive uncertainties of training examples based on DE. Then, we incorporate the resulting class imbalance measure $\PU_c$ into various methods during training the classifier.}
    \label{fig:arch}
\end{figure*}

In this section we incorporate our measure into a diverse set of imbalance mitigation methods following the taxonomy presented in~\cref{subsec:relatedwork}. \AB{As illustrated in~\cref{fig:arch}(a,b), our motivation is to place the decision boundary away from classes with higher \textsc{Class Uncertainty}, instead of classes with less samples as performed in the literature (e.g.,~\cite{khan2019striking,kim2020adjusting}).} \cref{fig:arch}\AB{(c)} visualizes the high-level scheme of our architecture to utilize predictive uncertainty to mitigate class imbalance. Specifically, having obtained class-wise uncertainty measures $\PU_c$ (\cref{eq:cw_uncertainty}) based on the predictive uncertainty estimates on the imbalanced training set, here we define $\alpha_c$ in~\cref{eq:resampling} for resampling methods; $w^{(i)}_c$ in~\cref{eq:lossweighting} for reweighting methods; $\Delta^{(i)}_c$ in~\cref{eq:softmax_margin} for margin-based methods; and finally the training strategy for multi-stage methods.
\textit{Uncertainty-based resampling (UBRs):} 
For resampling, we define $\alpha_c$ in~\cref{eq:resampling} for each class to promote or demote it based on its uncertainty.  The more uncertainty the examples from a class have, the more probability mass is. Since $\PU_c$ defined in~\cref{eq:cw_uncertainty} is already a valid probability distribution, we simply set $\alpha_c=\PU_c$.

\textit{Uncertainty-based reweighting (UBRw):} We define the weight of an example in class $c$, i.e., $w_c^{(i)}$ in~\cref{eq:lossweighting}, by multiplying $\PU_c$ by the total number of classes to ensure that the weights of the classes sum up to $|\mathcal{C}|$, which is the case once no reweighting is used and $w_c^{(i)}=1$ for all $c$. The resulting $w_c^{(i)}$ of the example $i$ of the class $c$ is then defined by
\begin{equation}
\label{eq:reweighting_unc}
    w_c^{(i)} = \PU_c \times |\mathcal{C}|.
\end{equation}
\textit{Uncertainty-based margins (UBM):} We incorporate margin-based uncertainties into LDAM~\cite{ldam} and DRO-LT Loss~\cite{Samuel_2021_ICCV}. Unlike UBRs and UBRm, for LDAM we use the unnormalized class uncertainties from~\cref{eq:cw_uncertainty_} considering that these methods offer normalizing the margins in their formulations and set the class margin $\Delta_c^{(i)}$ as:
\begin{equation}
    \Delta_c^{(i)} = \frac{\tau \times {\PURAW_{c}}}{\max_{c \in C}({\PURAW_{c}})},  
\end{equation}
with $\tau=0.5$ as in the original work. As for DRO-LT Loss~\cite{Samuel_2021_ICCV}, we again simply use $\Delta_c^{(i)} = \PU_c$.

\textit{Uncertainty-based multi-stage training:} Once we have the resampling and reweighting methods, incorporating them into two-stage training is straightforward: We train models without any imbalance mitigation strategy up to a certain epoch and then fine-tune them using UBRw or UBRs. 
\section{Experimental Results}
\label{section:exps}
\subsection{Implementation Details}
\label{subsec:imp_details}
\subsubsection{Obtaining Long-tailed Datasets} Similar to the literature~\cite{zhang2020tricks}, while obtaining CIFAR-10-LT and CIFAR-100-LT, we first estimate the number of examples for each class based on their indices as follows:
\begin{equation}
    N_c = \overbar{N} \times \frac{1}{\mathrm{IR}^{\frac{c}{|\mathcal{C}|-1}}},
\end{equation}
such that $\overbar{N}$ is the number of examples in each class in a balanced training set such as CIFAR-10 or CIFAR-100, $\mathrm{IR}$ is the imbalance ratio $\in \{50, 100\}$, and finally $|\mathcal{C}|$ is the total number of classes. Then, we randomly select $N_c$ examples from each class to obtain the long-tailed versions of the datasets.
\subsubsection{Analyzing \textbf{IF1(a)}} As suggested in~\cref{section:intro}, we use the class-wise accuracy of the na\"ive classifier on the test set as a proxy to the class imbalance.  We claim that these class-wise accuracies provide us with an indicator of the \textit{effects of the imbalance}. We use the na\"ive classifier as the mitigation methods are conventionally built to mitigate the effects of the class imbalance on this na\"ive classifier by modifying different steps in its pipeline (see~\cref{eq:lossweighting},~\cref{eq:resampling} and~\cref{eq:softmax_margin}). Note that these modifications require at least a proper ranking of the classes by the underlying imbalance measure such that they are promoted or demoted as intended, which is why we adopted the Spearman ranking correlation coefficient for quantification purposes. Furthermore, the accuracy/error on the \textit{test set} can not simply be employed as an imbalance measure, since it requires access to the test set during training, which is not possible. Note that this is also similar to the uncertainty calibration evaluation in which the methods are either tuned on the training set~\cite{devil,Norcal} or the validation set, and then evaluated against the accuracy of the test set~\cite{calibration,verifiedunccalibration}. We used CIFAR-100-LT with an IR of 50 for this analysis; enabling us to estimate $\rho$ for a larger population compared to CIFAR-10-LT. 
\subsubsection{Analyzing \textbf{IF1(b)}} Here, we simply increase the IR of the dataset, which corresponds to removing some examples from the classes in different numbers depending on IR. Particularly, as IR increases, the number of examples removed from the under-represented classes increase; making the resulting dataset more imbalanced. Therefore, considering that cardinality is one of the contributors of the class imbalance (but not the only one as previously discussed), this provides us with an easily controllable and applicable test bed to analyze this requirement. Specifically, we adopt CIFAR-10-LT with IR of $\{1, 2, 10, 20, 50\}$. Different from \textbf{IF1(a)}, here we prefer CIFAR-10-LT; allowing us to demonstrate the individual class behaviors in a way to be followed more easily.
\subsubsection{Analyzing \textbf{IF2}} Here, we aim to investigate the effect of duplicating examples, which basically corresponds to using oversampling to alleviate the impact of an imbalanced dataset. In order not to be limited for a single setting of the oversampling and cover the effect more thoroughly, we interpolate the sampling probability of the class $c$ denoted by $\alpha_c$ in~\cref{eq:resampling}. Following Peng et al.~\cite{Peng2020LargeScaleOD}, we achieve this by setting the sampling probability $\alpha_c$ as follows:
\begin{equation}
\alpha_c = (\alpha_c^{\mathcal{R}})^{\lambda} (\alpha_c^{\mathcal{N}})^{1-\lambda},    
\end{equation}
such that $\alpha_c^{\mathcal{R}}$ is the probability that an example will be sampled from class $c$ once oversampling is used (e.g., $\forall c, \alpha_c^{\mathcal{R}}=0.1$ for CIFAR-10 with 10 classes), $\alpha_c^{N}$ is the same probability but when no imbalance technique is used, and finally $\lambda \in [0,1]$ is the interpolation factor, which we term as the duplication strength. Note that when $\lambda = 0$, the dataset is in its original setting and $\lambda = 1$ implies that the dataset is over-sampled such that all classes have equal number of examples. We set the duplication strength to $\{0, 0.3, 0.5, 0.7, 1.0\}$ and similar to \textbf{IF1(b)}, we use CIFAR-10-LT to present the behavior of the classes clearly.
\subsubsection{Mitigating Class Imbalance}
\label{subsubsec:imp_mitigate}
In our implementation, unless otherwise specified, we exploit the common settings in the literature and implement our models upon the common benchmark as the official implementation of `Bag-of-Tricks'~\cite{zhang2020tricks} by keeping the settings of the baseline models, train a ResNet-32~\cite{ResNet} and adopt random horizontal flipping and random cropping for data augmentation. For optimization, we employ SGD with momentum using a batch size of 128 on a single GPU. We tune the learning rate of our method by using grid-search generally only in increments $0.1$ between $[0,0.7]$. We also note that since SVCI-20 is a new dataset, we also tuned the learning rates of the baseline methods such as the na\"ive classifier or LDAM to provide a fair comparison. All of the hyper-parameters including the learning rates that we tuned for our method will be made public with the release of our code upon acceptance.
\begin{table*}[hbt!]
\caption{Comparison of using \textsc{Class Uncertainty} instead of cardinality within  resampling, reweighting, margin adjustment, and multi-stage training methods.  UBRs: Uncertainty-based Resampling, UBRw: Uncertainty-based Reweighting, UBM: Uncertainty-based Margins. For each group, we directly take the result of competitive methods from the corresponding papers (indicated with $^*$) to show that our implementation is inline with the previously reported results. Among 20 different settings (4 datasets and 5 different set of methods), using our \textsc{Class Uncertainty} achieves the best results (in bold) in 16 different cases. Besides, the top scores in the table (underlined and bold) are obtained with \textsc{Class Uncertainty} as well.}
\label{table:results}
\centering
    \begin{tabular}{ |c|c|c c|c c|}
    \hline \rowcolor{Gray}
    &Dataset & \multicolumn{2}{c|}{CIFAR-10-LT} & \multicolumn{2}{c|}{CIFAR-100-LT}\\
    \hline\rowcolor{LightGray}
    &Imbalance Ratio (IR) & 50 & 100 & 50 & 100\\
    \hline \hline
    &Na\"ive Classifier& 24.26±0.35 & 30.18±0.52 & 57.14±0.08 & 61.48±0.40 \\ \hline \hline 
    \multirow{5}{*}{\rotatebox{90}{\small{Resampling}}} &CB Resampling~\cite{kang2019decoupling} & 23.33±0.14 & 30.33±0.30 & 61.13±0.52 & 66.34±0.30 \\
    &PB Resampling$^{*}$~\cite{zhang2020tricks} & 25.03 & 32.91 & 57.09 & 61.41 \\
    &PB Resampling~\cite{kang2019decoupling} & 23.86±0.36 & 29.62±0.49 & 56.87±0.13 & 60.40±0.57 \\
    \cline{2-6}
    &UBRs \textbf{(Ours)} & 23.32±0.11 & 29.09±0.37 & 66.12±0.73 & 72.53±0.74\\
    &PB UBRs \textbf{(Ours)} & \textbf{22.61±0.70} & \textbf{28.67±0.91} & \textbf{55.48±0.08} & \textbf{59.28±0.47}\\
    \hline \hline 
    \multirow{5}{*}{\rotatebox{90}{\small{Reweighting}}}&
    Focal Loss~\cite{focal} & 23.91±0.36 & 30.49±0.68 & 57.50±0.07 & 61.32±0.58 \\
    &Class-balanced Loss$^{*}$~\cite{cui2019classbalancedloss} & \textbf{20.73} & \textbf{25.43} & 54.68 & 60.40 \\
    &Class-balanced Focal Loss~\cite{cui2019classbalancedloss} & 21.29±0.39 & 26.64±0.59 & 57.60±0.48 & 61.41±0.35 \\
    \cline{2-6}
    &UBRw \textbf{(Ours)} & 22.51±0.61 & 26.72±0.59 & 55.23±0.53 & 59.23±0.42 \\
    &UBRw Focal Loss \textbf{(Ours)} & 21.63±0.42 & 26.54±0.28 & \textbf{53.85±0.71} & \textbf{58.33±0.49} \\ \hline \hline
    \multirow{7}{*}{\rotatebox{90}{\small{Margin Adj.}}}
    &LDAM~\cite{ldam}& 23.31±0.11 & 27.95±0.20 & 57.72±0.27 & 61.22±0.10 \\
    &LDAM+Reweighting$^*$~\cite{ldam}& N/A & 22.97 & N/A & 57.96 \\
    &LDAM+Reweighting~\cite{ldam} & 20.32±0.02 & 23.90±0.28 & 54.27±0.43 & 57.98±0.20 \\
    &DRO-LT~\cite{Samuel_2021_ICCV} &14.63±0.10 & 17.99±0.11 & 46.75±0.08 & 51.85±0.13\\
    \cline{2-6}
    &UBM LDAM \textbf{(Ours)} & 22.74±0.23 & 27.31±0.39 & 55.50±0.49 & 59.19±0.07 \\
    &UBM LDAM+Reweighting \textbf{(Ours)} & 20.20±0.29 & 21.80±0.33 & 50.40±0.10 & 54.55±0.14 \\
    &UBM DRO (\textbf{Ours)} & \textbf{\underline{14.54±0.25}} & \textbf{\underline{17.68±0.31}} & \textbf{\underline{46.54±0.11}} & \textbf{\underline{51.65±0.08}}
    \\[0.25em]
    \hline \hline
    \multirow{13}{*}{\rotatebox{90}{\small{Multi-stage Methods}}}&\emph{\small{Two-stage Resampling:}} & & & & \\
    &CB Resampling~\cite{kang2019decoupling} & 21.47±0.36 & 27.00±0.21 & 54.39±0.15 & 58.83±0.14 \\
    &PB Resampling$^*$~\cite{zhang2020tricks} & 24.58 & 33.48 & 56.93 & 61.35 \\
    &PB Resampling~\cite{kang2019decoupling} & 22.56±0.10 & 28.99±0.72 & 54.96±0.26 & 59.27±0.30 \\
    &CAM-based Resampling$^*$~\cite{zhang2020tricks} & \textbf{18.66} & 22.62 & 53.56 & 57.70 \\
    \cline{2-6} 
    &UBRs  \textbf{(Ours)} &19.06±0.14 & \textbf{22.50±0.14} & \textbf{51.82±0.44} & \textbf{56.34±0.46} \\
    \cline{2-6}
    &\emph{\small{Two-stage Reweighting:}} & & & & \\
    &Cost-sensitive Cross Entropy~\cite{csce} & 21.03±0.46 & 26.11±0.22 & 54.41±0.25 & 58.78±0.13 \\
    &Focal Loss~\cite{focal} & 22.03±0.27 & 28.51±0.68 & 57.23±0.05 & 61.70±0.30 \\
    &Class-balanced Focal Loss$^*$~\cite{zhang2020tricks} & 20.81 & 25.31 & 54.57 & 58.92 \\
    &Class-balanced Focal Loss~\cite{cui2019classbalancedloss} & \textbf{20.45±0.27} & 26.08±0.34 & 57.09±0.55 & 59.90±0.19 \\
    \cline{2-6}
    &UBRw \textbf{(Ours)} & 22.21±0.35 & 26.41±0.14 & \textbf{52.90±0.06} & \textbf{57.25±0.17} \\
    &UBRw Focal Loss \textbf{(Ours)} & 22.22±0.22 & \textbf{25.71±0.21} & 53.27±0.08 & 57.89±0.13 \\
    \hline
    \end{tabular}
\end{table*}
\subsection{Results on Long-tailed Datasets}

\subsubsection{Datasets and Performance Measure} Following the literature~\cite{zhang2020tricks, Samuel_2021_ICCV, ldam, cui2019classbalancedloss}, we use the CIFAR-10-LT and CIFAR-100-LT datasets. These are obtained by systematically reducing the number of examples for certain classes in CIFAR-10 and CIFAR-100~\cite{cifar} so that the datasets follow a long-tailed distribution (see~\cref{subsec:imp_details} for details). Specifically, we used IRs of 50 and 100 following Zhang et al.~\cite{zhang2020tricks}. For testing, we used the original (balanced) test sets of CIFAR-10 and CIFAR-100 with 10K examples and reported the top-1 error.

\subsubsection{Resampling}
We first compare our Uncertainty-based Resampling (UBRs) with CB resampling and PB resampling~\cite{kang2019decoupling}, both of which rely on class cardinality as discussed in~\cref{subsec:relatedwork}. Similarly, we train the classifier by (i) simply using the class-wise sampling probabilities obtained through \textsc{Class Uncertainty} (UBRs); and (ii) inspired by PB, gradually changing the weights towards our uncertainty-based weights (PB UBRs). \cref{table:results} suggests that UBRs performs better than both methods on CIFAR-10-LT but not on CIFAR-100-LT, which is a more challenging dataset. For example, while the class with minimum cardinality has 50 training examples for CIFAR-10-LT, it is only 5 for CIFAR-100-LT. Note that using our PB UBRs consistently outperforms all resampling methods. For example, \textit{PB UBRs decreases the top-1 error rate of PB by $~1.5\%$ on CIFAR-100-LT on both IRs}. These results suggest that our PB UBRs is more effective than existing resampling methods relying on class cardinality.

\subsubsection{Loss Reweighting}
Here, we consider two common baselines: (i) Focal Loss~\cite{focal}, as a hardness-based method ~\cite{oksuz2020imbalance}, and (ii) Class-balanced Loss, a cardinality-based method using the ``effective'' number of examples in each class~\cite{cui2019classbalancedloss}. Note that Class-balanced Loss is shown to be a stronger approach compared to directly relying on the number of examples~\cite{cui2019classbalancedloss,zhang2020tricks}. \cref{table:results} displays that:
\begin{itemize}
    \item Weighting the na\"ive classifier using our \textsc{Class Uncertainty} (UBRw) outperforms Focal Loss and consistently improves performance (up to $3.77 \%$).
    \item Incorporating \textsc{Class Uncertainty} into Focal Loss (similar to Class-balanced Loss) outperforms the Class-Balanced Loss by up to more than $2 \%$ on CIFAR-100-LT ($60.40$ vs. $58.33$ with IR of 100).
\end{itemize}
These results demonstrate the effectiveness of our approach in loss reweighting methods.
\subsubsection{Margin Adjustment}
We replace the cardinality-based margins with our \textsc{Class Uncertainty} in three different methods: (i) LDAM~\cite{ldam}; (ii) LDAM combined by reweighting~\cite{ldam} as a more competitive approach; and (iii) DRO-LT Loss~\cite{Samuel_2021_ICCV} with cardinality-based margins. We observe in~\cref{table:results} that:
\begin{itemize}
\item Using margins based on \textsc{Class Uncertainty} in LDAM (UBM LDAM) consistently outperforms LDAM for both settings. Specifically, our gains are significant on CIFAR-100-LT, a situation similar to what we have observed for resampling and loss weighting methods. \textit{For example, UBM LDAM with reweighting improves its counterpart between $3.5-4 \%$ top-1 error on CIFAR-100-LT.}
\item As for DRO-LT Loss, UBM improves the baseline consistently across all four settings; \textit{reaching the top performance in~\cref{table:results}.}
\end{itemize}
\subsubsection{Multi-stage Training}
We incorporate our imbalance measure into three multi-stage training methods following Zhang et al.~\cite{zhang2020tricks}. That is, we train a baseline model without any imbalance mitigation for the first 160 epochs, then use UBRs, UBRw, and UBRw Focal Loss with our \textsc{Class Uncertainty} in the second stage. In addition to the baselines from the previous sections, we also compare our method with CAM-based sampling using class activation maps for resampling~\cite{zhang2020tricks} and CSCE~\cite{csce} based on class cardinalities. \cref{table:results} shows that:
\begin{itemize}
\item Over eight different dataset and IR combinations, our methods perform the best on six of them, with the other two being relatively on-par with common baselines.
\item Similarly to what we observed in resampling and loss reweighting, using \textsc{Class Uncertainty} in the second stage also improves the performance, especially for CIFAR-100-LT. For example, UBRs performs around $1.5-2.0\%$ top-1 error better than CAM-based resampling, its closest counterpart.
\item Regarding loss weighting, the na\"ive UBRw performs the best in CIFAR-100-LT with arguably similar results on CIFAR-10-LT.
\end{itemize}
\begin{figure*}[hbt!]
        \centering
        \subfloat[Effect of Number of Examples ($N_c$)]{\includegraphics[width=0.35\textwidth]{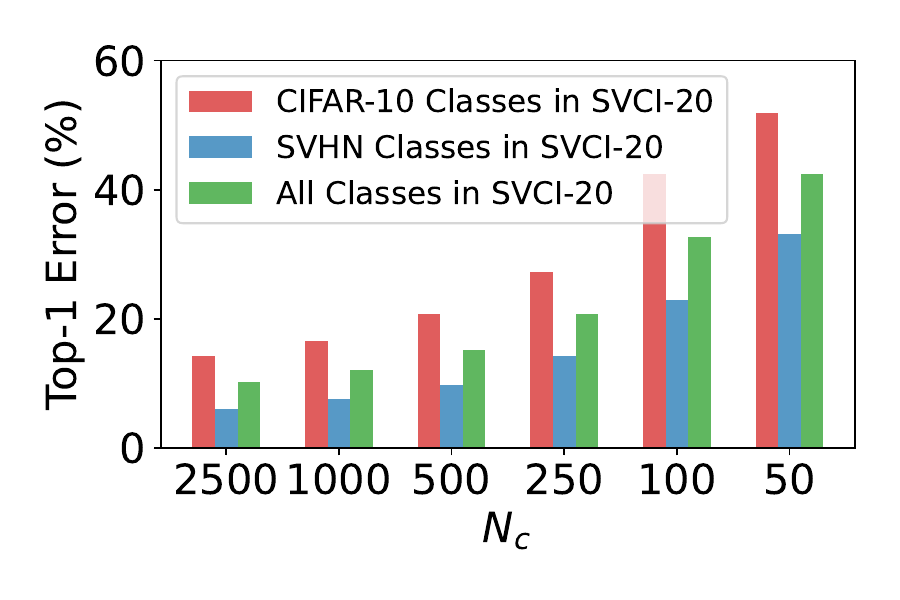}}
        \subfloat[$\CC_c$ vs. $\PU_c$ on SVCI-20 with $N_c=100$]{\includegraphics[width=0.35\textwidth]{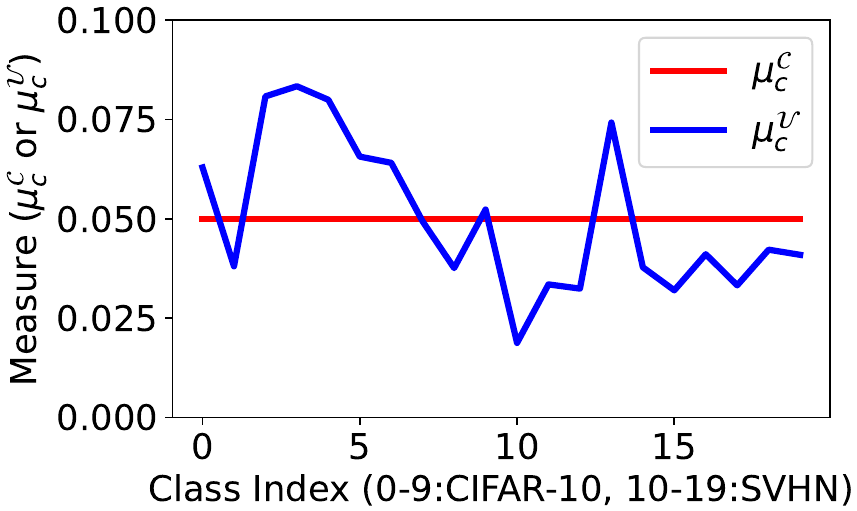}}
        \caption{
        \textbf{(a)} As the number of examples from each class ($N_c$) decreases, the error gap between CIFAR-10 and SVHN classes in SVCI-20 dataset increases for Na\"ive Classifier; implying each setting corresponds to a different semantic imbalance level.        
        \textbf{(b)} While \textsc{Class Cardinality Imbalance Measure} ($\CC_c$) does not prioritize any class on SVCI-20, our \textsc{Class Uncertainty} ($\PU_c$) differentiates between them. Generally, $\PU_c$ assigns more priority to CIFAR-10 classes (class index less than or equal to 9) as the harder classes. }
        \label{fig:svci}
\end{figure*}
\begin{table}[hbt!]
    \setlength{\tabcolsep}{0.3em}
    \caption{Performance gain of using \textsc{Class Uncertainty}. While using class cardinality-based methods generally corresponds to the Na\"ive classifier, our \textsc{Class Uncertainty} approach is applicable to any existing class imbalance mitigation methods; providing notable gains on our semantically-imbalanced settings. Note that the performances on the first four rows are the same as class cardinalities are the same. \label{tab:results_semantic}}
    \centering
    \begin{tabular}{ |c|c c c|}
    \hline \rowcolor{Gray}
    Dataset & \multicolumn{3}{c|}{SVCI-20}\\
    \hline\rowcolor{LightGray}
    $N_c$ & 250 & 100 & 50 \\
    \hline \hline
    Na\"ive Classifier& 22.04 ± 0.65 & 33.56 ± 1.40 & 46.17 ± 3.10 \\
    \hline \hline
    \emph{\small{Resamp. \& Reweight.:}} & & & \\
    PB Rs~\cite{kang2019decoupling}& 22.04±0.65 & 33.56±1.40 & 46.17±3.10 \\
    Class-balanced Loss~\cite{cui2019classbalancedloss}& 22.04±0.65 & 33.56±1.40 & 46.17±3.10  \\
    Two-stage CSCE Rw~\cite{csce}& 22.04±0.65 & 33.56±1.40 & 46.17±3.10  \\
    Focal Loss~\cite{focal} & 21.65±0.04 & 33.61±0.97 & 50.37±0.48  \\
    \hline
    PB UBRs \textbf{(Ours)} &\textbf{21.38±0.34}& \textbf{32.78±0.68}& \textbf{45.31±1.30} \\
    \hline \hline 
    \emph{\small{Margin Adjustment:}}  & & & \\
    LDAM~\cite{ldam}&20.72±0.80 &\textbf{\underline{30.90±0.39}}& 43.36±0.57  \\
    \hline
    UBM LDAM \textbf{(Ours)} & \textbf{\underline{19.87±0.47}}& 31.20±0.09 & \textbf{\underline{42.17±0.39}} 
    \\
    \hline
    \end{tabular}
\end{table}
We conducted ablation studies to explore the flexibility of our approach by combining \textsc{Class Uncertainty} with the effective number of examples~\cite{cui2019classbalancedloss} or using DUQ~\cite{duq} for more reliable uncertainties. \AB{We also provided additional results on a real-world dataset, i.e. Places-LT~\cite{liu2019oltr}, where we improve over the baseline.} These findings suggest potential for further improvements (see the supplementary material).
\subsection{Results on Semantically-Imbalanced Datasets}
\label{subsec:semanticalexp}
\subsubsection{Dataset Curation} Here, we introduce SVCI-20, a balanced dataset in terms of cardinality but semantically-imbalanced dataset, to test our models more thoroughly and foster research in this direction. Specifically, SVCI-20 combines  CIFAR-10~\cite{cifar} including 10 classes with SVHN~\cite{SVHN} which consists of 10 digits taken from house numbers; totaling to 20 classes. Arguably, SVHN is an `easier' dataset than CIFAR-10 considering that: (i) Existing work reports a higher accuracy for SVHN, e.g., DenseNet~\cite{DenseNet} with a depth of 40 obtains 7.00 top-1 error on CIFAR-10, while it has 1.79 top-1 error on SVHN; and (ii) from a model-free perspective, the intrinsic dimension of SVHN is $1.5 \times$ smaller than that of CIFAR-10~\cite{intrinsicdim}. Therefore, combining these datasets into SVCI-20 increases the semantic gap among classes and using uniform class cardinalities removes the impact of cardinality imbalance, both of which are suitable for our purpose. We measure the performance using the average top-1 error of the classes.

\subsubsection{Effect of Class Cardinality on Dataset Curation} Here, we seek to construct SVCI-20 by setting $N_c$ (the number of examples taken from classes) properly.
Note from~\cref{fig:svci}(a) that as $N_c$ decreases: (i) the Top-1 error increases since the training data decrease for all classes; and more importantly, (ii) the error gap between the CIFAR-10 and SVHN classes in SVCI-20 also increases. As an example, when $N_c=1000$, the gap is $\sim 9 \%$, while it is around $20 \%$ for $N_c=50$. Based on these observations, we randomly sampled $250$, $100$ and $50$ training examples (i.e., $N_c$s for all $c \in \mathcal{C}$) to further analyze different levels of semantic imbalance, as we did with different IRs in long-tailed datasets.

\subsubsection{Class Cardinality on SVCI-20} As all classes have equal number of training examples, cardinality is indistinctive between classes (the red curve in~\cref{fig:svci}(b)). Consequently, except for Focal Loss and LDAM, the methods reduce to the Na\"ive Classifier; effectively ignoring this type of class imbalance (\cref{tab:results_semantic}).

\subsubsection{\textsc{Class Uncertainty} on SVCI-20} On the SVCI-20 dataset, \textsc{Class Uncertainty} is expected to be more distinctive between classes compared to Class Cardinality since it considers other aspects. \cref{fig:svci}(b) shows that \textsc{Class Uncertainty} for the CIFAR-10 classes is generally higher than those for the SVHN classes; making our approach different from class cardinality. Consequently, (i) our PB UBRs achieves the best results among resampling and rewighting methods; and (ii) UBM LDAM outperforms LDAM in two out of three cases $\sim 1 \%$ while being on par for $N_c=100$ (\cref{tab:results_semantic}).
\section{Conclusion}
\label{sec:discuss}
In this work, we demonstrate that \textsc{Class Cardinality Imbalance Measure}, as the de facto measure of class imbalance, has some limitations, and our extensive analyses and experiments show that \textsc{Class Uncertainty} is a better alternative. 
Our approach offers two important benefits. For the former, our measure \textsc{Class Uncertainty} can be considered as easy-to-integrate for imbalance mitigation techniques, since we only tune the learning rate while incorporating our method, and in very rare cases, such as LDAM, search for the method-specific hyperparameters. As for the latter, our perspective to devise new measures is promising as we obtain better performance on our new test bed comprising of a semantically-imbalanced dataset with equal number of examples such that cardinality-based approaches would fail to capture the difference across classes.

It is also worth noting that one related work~\cite{khan2019striking} employs epistemic uncertainty to mitigate imbalance, but in a specific way that fits in their proposed loss function. Instead, we consider using predictive uncertainty to mitigate class imbalance from a broader perspective by (i) proposing a new measure based on predictive uncertainty, (ii) extensively analyzing, and (iii) using the measure with ten imbalance mitigation approaches with different characteristics in imbalance setups.

\AB{\textbf{Limitations.} In this study, we use Deep Ensembles~\cite{de}, a prominent method for measuring uncertainty, in a two-stage training pipeline to decouple the analysis and experimental evidence we provide. Noting that our main focus is the analysis of class uncertainty as an imbalance measure, the adopted approach comes with high computational complexity in terms of memory and time. Moreover, the reliability of predictive uncertainty is based on accurate quantification, which recent studies have set an upper limit. Although prior research has touched on the concept of ``ideal uncertainty'' in Bayes-optimal predictors~\cite{Jain2021DEUPDE}, there is much more to explore in defining this ``ideal'' state. Understanding the impact of this ideal uncertainty on class imbalance is a crucial research question originating from this perspective, which remains outside the scope of this work.}

 \bibliographystyle{IEEEtran} 
 \bibliography{ms}

\end{document}


\title{Supplementary Material for \\Class Uncertainty: \\A Measure to Mitigate Class Imbalance}

\author{Zeynep Sonat Baltaci$^{1,2}$, Kemal Oksuz$^{3}$, Selim Kuzucu$^{1}$, Kivanc Tezoren$^{1}$, Berkin Kerim Konar$^{1}$, \\Alpay Ozkan$^{1}$, Emre Akbas$^{1,4,\dagger}$, Sinan Kalkan$^{1,4,\dagger}$\\
  $^1$Dept. of Computer Engineering, METU, Ankara, Turkey \\
  $^2$LIGM, Ecole des Ponts, Univ Gustave Eiffel, CNRS, Marne-la-Vallée, France \\
  $^3$Five AI Ltd., United Kingdom \\
  $^4$Center for Robotics and Artificial Intelligence (ROMER), METU, Ankara, Turkey \\
  sonat.baltaci@enpc.fr,  kemal.oksuz@five.ai, \\
  \{selim.kuzucu, kivanc.tezoren, berkin.konar, alpay.ozkan, eakbas, skalkan\}@metu.edu.tr
    \thanks{Manuscript received October 3, 2024.}}

\markboth{Under Review}%
{Supplementary Material for Class Uncertainty: A Measure to Mitigate Class Imbalance}

\maketitle
\def\thefootnote{$\dagger$}\footnotetext{Equal contribution for senior authorship.}\def\thefootnote{\arabic{footnote}}
\section{Further Analyses}
\subsection{Using Different Uncertainty Estimation Methods}
\label{sec:sup1a}
\begin{figure}[hbt!]
        \centering
        \includegraphics[width=0.3\textwidth]{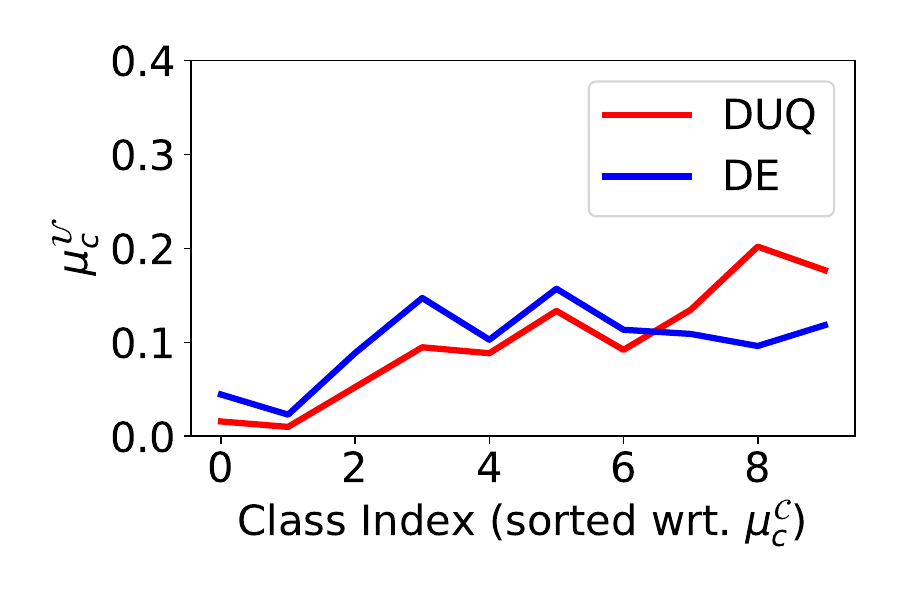}
        \caption{$\PU_c$ for DE vs. DUQ. DE and DUQ results in a similar trend in \textsc{Class Uncertainty} ($PU_c$), while the uncertainties from DUQ promoting under-represented classes slightly more.}
        \label{fig:deduq}
\end{figure}

The uncertainty estimation method plays an important role for our \textsc{Class Uncertainty}. Although we employ DE~\cite{de} to quantify predictive uncertainty due to its robustness over varying datasets, here we show that more reliable uncertainty quantification methods may bring an improvement. As an initial study, we employ DUQ~\cite{duq}, which arguably yields more reliable uncertainties compared to DE on CIFAR-10~\cite{cifar}. \cref{fig:deduq} compares the \textsc{Class Uncertainties} obtained from DE and DUQ, indicating a similar trend between the two, with DUQ promoting under-represented classes more. \cref{tab:deduq} shows that the use of DUQ with FL performs better on CIFAR-10-LT~\cite{cui2019classbalancedloss}. Note that we only provide results on CIFAR-10 since DUQ has not been applied to CIFAR-100 yet, and scaling DUQ to larger datasets is beyond the scope of our work. 

\AB{Uncertainty quantification is an active research area where the literature has produced many alternative approaches with different advantages and disadvantages (see, e.g.,~\cite{abdar2021review,Gawlikowski2021ASO}). Although the results in~\cref{fig:deduq} and~\cref{tab:deduq} suggest that there is room for improvement with more reliable uncertainty estimation techniques, using DE-based uncertainties does work better in some cases and provides comparable gains over the baselines. In light of this, we prefer to exclude a more comprehensive comparison using alternative uncertainty quantification approaches to preserve the focus of the paper.}
\begin{table}[hbt!]
    \centering
        \caption{Comparing different uncertainty quantification methods. Combining the \textsc{Class Uncertainties} of DUQ with Focal Loss yields the best performance. Compared to the Na\"ive Classifier, the improvement is between $3$ to $4$ top-1 error.}
        \label{tab:deduq}
        \begin{tabular}{|c|c| c c|}
        \hline 
         \multirow{ 2}{*}{Method}&\multirow{ 2}{*}{Unc. Qua.}& \multicolumn{2}{c|}{Imbalance Ratio (IR)} \\
        \cline{3-4} 
         && 50 & 100 \\
        \hline \hline
        Na\"ive Cls. &N/A& 24.26±0.35 & 30.18±0.52 \\
        \hline
        UBRw & DE & 22.51±0.61 & 26.72±0.59 \\
        UBRw & DUQ & 22.84±0.36 & 27.60±0.89 \\
        \hline
        UBRw Focal Loss  & DE & 21.63±0.42 & 26.54±0.28 \\
        UBRw Focal Loss  & DUQ &\textbf{20.90±0.17} & \textbf{25.22±0.87} \\
        \hline
        \end{tabular}
\end{table}
\subsection{Combining \textsc{Class Uncertainty} and Cardinality}
We finally investigate the benefit of combining \textsc{Class Uncertainty} with class cardinality. Specifically, we obtain the weights in Eq. (4) (see the main article) by combining our UBRw Focal Loss~\cite{focal} and Class-balanced Focal Loss (CB FL)~\cite{cui2019classbalancedloss} simply by,
\begin{equation}
w_c^{(i)} = \gamma w_c^{\mathcal(U)} + (1- \gamma) w_c^{\mathcal(C)},
\end{equation}
where $w_c^{\mathcal(U)}$ and $w_c^{\mathcal(C)}$ are the weights estimated by our UBRw FL and CB FL with $\gamma \in [0,1]$ being the weighting coefficient. \cref{tab:cb_ubrw} shows on both CIFAR-10-LT and CIFAR-100-LT that training benefits from this combination. These results suggest the potential benefit of combining multiple imbalance measures.
\begin{table}[hbt!]
    \centering
        \caption{Performance gain of using \textsc{Class Uncertainty} compared to Na\"ive classifier and Focal Loss. Combining \textsc{Class Uncertainties} with the effective number of samples proposed by Cui et al.~\cite{cui2019classbalancedloss} promises further improvement; suggesting better measures to capture class imbalance can be devised with a more thorough investigation on combining different measures. $\gamma= 0.0$ and $\gamma= 1.0$ corresponds to CB FL and UBRw FL, respectively.} 
        \label{tab:cb_ubrw}
    \begin{tabular}{ |c|c c|c c|}
    \hline \rowcolor{Gray}
    Dataset & \multicolumn{2}{c|}{CIFAR-10-LT} & \multicolumn{2}{c|}{CIFAR-100-LT}\\
    \hline \rowcolor{LightGray}
    IR & 50 & 100 & 50 & 100\\
    \hline
    \hline
    $\gamma= 0.0$ & 21.29±0.39 & 26.64±0.59 & 57.60±0.48 & 61.41±0.35 \\
    
    $\gamma= 0.3$ & 20.22±0.20 & 25.39±0.09 & 54.24±0.57 & 59.61±0.49 \\
    
    $\gamma= 0.5$ & \textbf{20.51±0.37} & \textbf{25.27±0.47} & 53.73±0.54 & 58.30±0.30 \\
    
    $\gamma= 0.7$ & 20.74±0.51 & 25.82±0.20 & \textbf{53.33±0.23} & \textbf{58.26±0.29} \\
    
    $\gamma= 1.0$ & 21.63±0.42 & 26.54±0.28 & 53.85±0.71 & 58.33±0.49 \\
    \hline
    \end{tabular}
\end{table}
\subsection{Using Different Uncertainty Types as a Class Imbalance Measure}
\begin{table*}[ht!]
    \caption{\AB{Empirical comparison of predictive and epistemic uncertainty: Top-1 error rates of predictive and epistemic uncertainty as measures on CIFAR-10-LT.}}
    \centering
    \resizebox{\textwidth}{!}{
    \begin{tabular}{ |M{3cm}|M{2cm} M{2cm}|M{2cm} M{2cm}|M{2cm} M{2cm}|M{2cm} M{2cm}|}
    \hline
    \rowcolor{Gray}
    Dataset & \multicolumn{4}{|c|}{CIFAR-10-LT} & \multicolumn{4}{|c|}{CIFAR-100-LT} \\
    \hline
     \rowcolor{LightGray}
     Imbalance Ratio & 50 & 100 & 50 & 100 & 50 & 100 & 50 & 100 \\
     \hline
     Uncertainty Type & \multicolumn{2}{|c|}{Predictive Unc.} & \multicolumn{2}{c|}{Epistemic Unc.} & \multicolumn{2}{|c|}{Predictive Unc.} & \multicolumn{2}{c|}{Epistemic Unc.}  \\
    \hline \hline
    PB UBRs & 22.61±0.70 & 28.67±0.91 & \textbf{21.84±0.57} & \textbf{27.92±0.61} & 55.48±0.08 & 59.28±0.47 & \textbf{55.08±0.67} & \textbf{58.95±0.39} \\
    \hline
    UBRw & 22.51±0.61 & \textbf{26.72±0.59} & \textbf{21.75±0.13} & 28.06±1.06 & \textbf{55.23±0.53} & \textbf{59.23±0.42} & 55.62±0.20 & 59.48±0.57 \\
    \hline
    UBRw Focal Loss & 21.63±0.42 & 26.54±0.28 & \textbf{21.59±0.88} & \textbf{25.57±0.61} & 53.85±0.71 & \textbf{58.33±0.49} & \textbf{53.09±0.61} & 59.03±1.36 \\
    \hline
    \end{tabular}}
    \label{tab:epis-cifar10}
\end{table*}
\AB{Epistemic uncertainty represents the lack of information within a data space, which inherently reflects the cardinality imbalance among classes. Starting from this point of view, we compare predictive and epistemic uncertainty as representatives of class imbalance empirically. We employ mutual information for epistemic uncertainty as in~\cite{Gal2016UncertaintyID}. \cref{tab:epis-cifar10} shows the performance of UBRw and UBRs on CIFAR-LT datasets. Although there is not much of a performance gap between predictive and epistemic uncertainty utilized as an imbalance measure, the more the complexity increases regarding the number of classes and the class cardinality difference, the better predictive uncertainty performs. This constructs empirical support for predictive uncertainty as an imbalance measure stating the fact that it also inherently reflects aleatoric uncertainty.}
\subsection{Comparison of Uncertainty Types for Capturing Class Uncertainty}
\begin{figure}
    \centering
     \subfloat{\includegraphics[width=0.5\textwidth, trim={4cm 7cm 0 3cm}, clip]{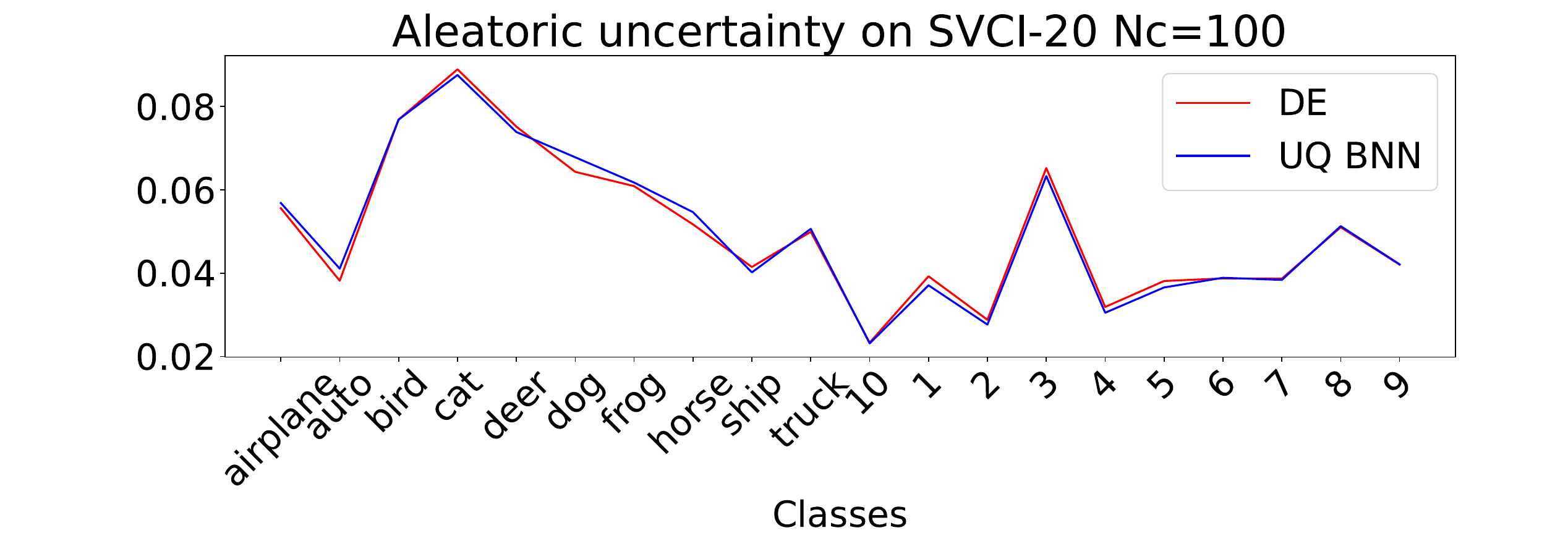}}\vspace{1mm}
     \subfloat{\includegraphics[width=0.5\textwidth, trim={4cm 7cm 0 3cm}, clip]{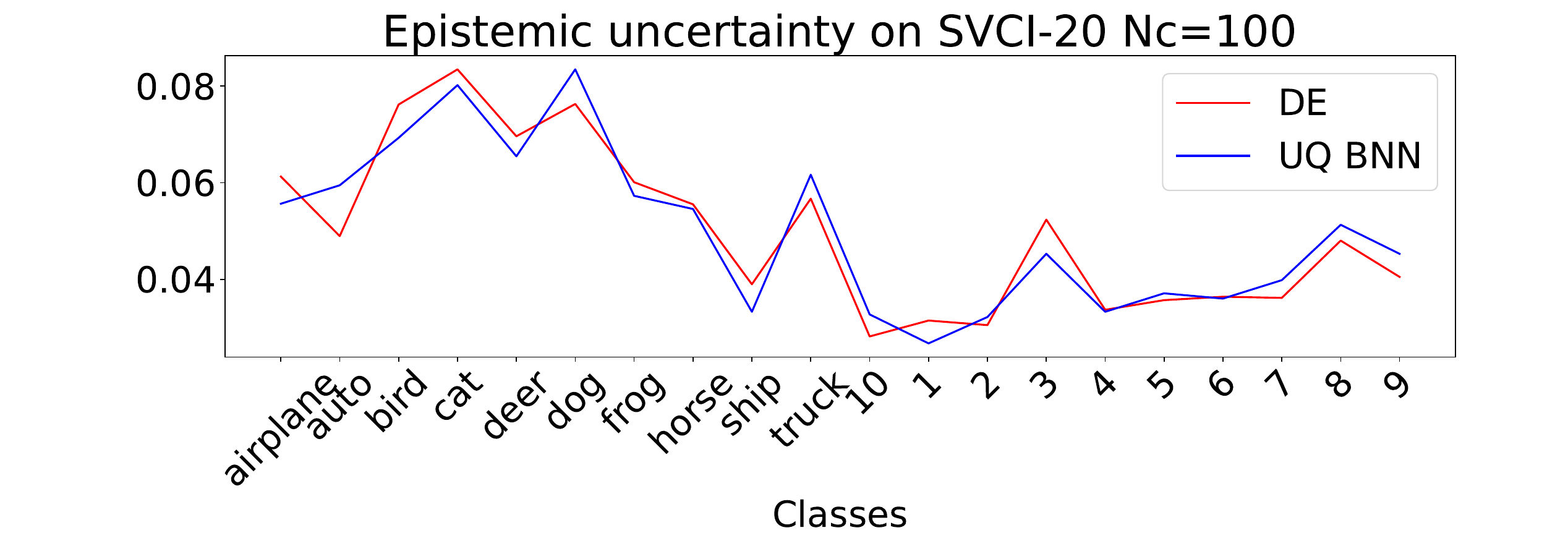}}\vspace{1mm}
      \subfloat{\includegraphics[width=0.5\textwidth, trim={4cm 7cm 0 3cm}, clip]{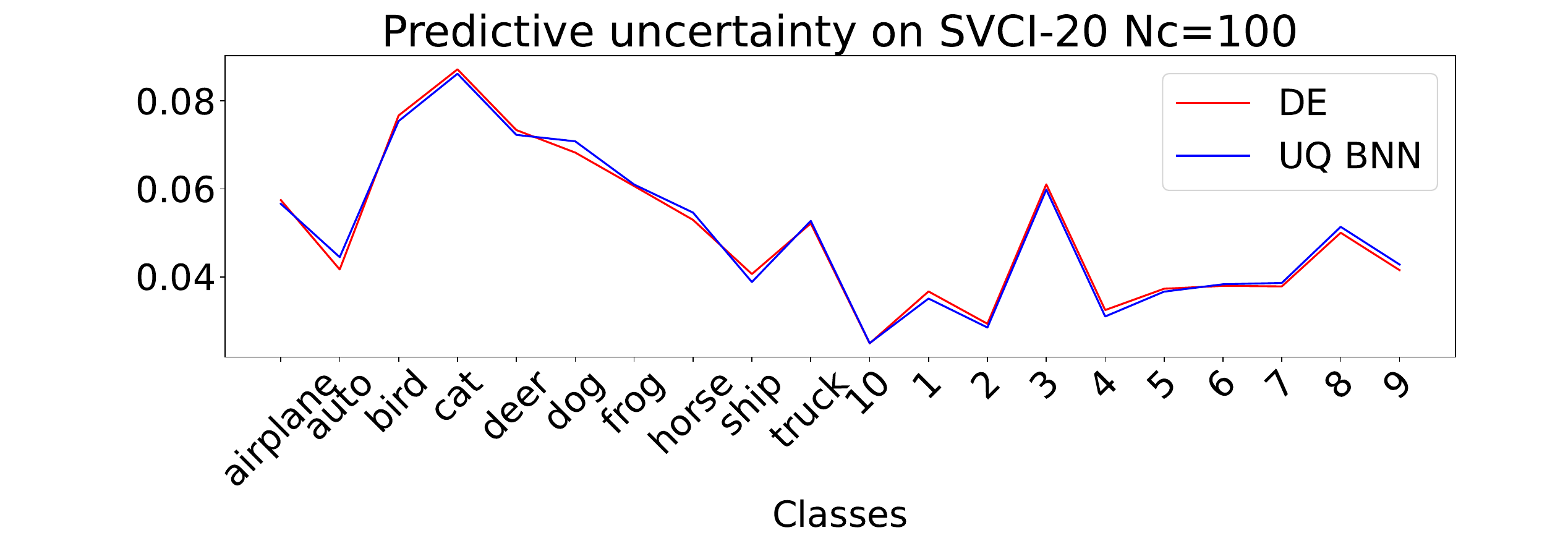}}
    \caption{Aleatoric, epistemic and predictive uncertainties of classes in the SVCI-20 dataset, quantified with different methodologies.}
    \label{fig:unc-types-svci}
\end{figure}

\AB{We recognize that conducting a thorough and theoretical examination of both epistemic and aleatoric uncertainty, and their implications on classifier performance would yield valuable insights into data and model interpretation. As a base comparison, we examine epistemic, aleatoric, and predictive uncertainty calculated with the algorithm proposed by Kwon et al.~\cite{kwon2020uq} on the SVCI-20 dataset with 100 samples in each class along with Deep Ensembles. We exploit the predictions of the ensemble members as MC samples in the algorithm. We perform this analysis on SVCI-20 to exclude the possible effects of cardinality. As depicted in~\cref{fig:unc-types-svci}, there is a high correlation between the two quantification methods. Moreover, aleatoric uncertainty represents the majority of quantified uncertainty, which is consistent with the complexity of CIFAR-10 compared to SVHN.}
\AB{Thus, within the confines of this study, our focus remains on predictive uncertainty determined through entropy, inherently encapsulating both epistemic (model-related) and aleatoric (data-related) uncertainty in imbalance cases.}
\subsection{Experiments on Different Architectures}
\AB{We employed the Swin Tiny transformer~\cite{liu2021Swin} as a feature extractor rather than ResNet-32 to investigate the robustness of our approach and potential enhancements associated with an architecture with higher complexity and context capability. Unlike how we used ResNet-32, the utilization of a pretrained model is necessary due to the substantial computational complexity in training a Swin transformer from scratch. Due to this computational complexity, we only compare one-stage mitigation methods. As presented in~\cref{tab:swin}, uncertainty-based mitigation methods not only improve baseline but perform better than class cardinality-based counterparts. Although our preliminary findings show promise, a comprehensive examination is crucial to discern the implications of acquired features, evaluate the performance of a context-aware model in the context of imbalanced data, and identify potential performance enhancements with the mitigation methods proposed, thus allowing a comprehensive and definitive conclusion.}
\begin{table}[htbp!]
    \centering
    \caption{\AB{Top-1 error rate of mitigation methods with Swin Transformer as a feature extractor on CIFAR-10-LT and CIFAR-100-LT with IR 50.}}
    \resizebox{\columnwidth}{!}{
    \begin{tabular}{|c|c|c|}
        \hline
        \rowcolor{Gray}
         Dataset & CIFAR-10-LT & CIFAR-100-LT \\
         \hline
         Na\"ive Cls. & 9.71±0.12 & 31.14±0.06 \\
         \hline \hline
         \emph{\small{Resampling}} & & \\
         CB Resampling~\cite{kang2019decoupling} & 6.60±0.07 & 28.88±0.20 \\
         PB Resampling~\cite{kang2019decoupling} & 6.71±0.07 & 28.47±0.10 \\
         \hline
         UBRs \textbf{(Ours)} & 6.35±0.03 & 29.35±0.56 \\
         PB UBRs \textbf{(Ours)} &\textbf{6.26±0.21} & \textbf{27.52±0.23} \\ 
         \hline \hline
         \emph{\small{Reweighting}} & & \\
         Class-balanced Focal Loss~\cite{cui2019classbalancedloss} & 6.67±0.14 & \textbf{28.13±0.16} \\
         \hline
         UBRw Focal Loss \textbf{(Ours)} & \textbf{6.50±0.09} & 28.72±0.21 \\
         \hline
    \end{tabular}}
    \label{tab:swin}
\end{table}
\subsection{Statistical Significance Tests}
\begin{table}[htbp!]
    \caption{\AB{$\epsilon_{min}(\downarrow)$ values of ASO test. The results which our method outperforms the counterpart are in bold.}}
    \label{tab:sig}
\centering
\resizebox{\columnwidth}{!}{
    \begin{tabular}{|c c|c c|c c|}
    \hline
    \rowcolor{Gray}
    \multicolumn{ 2}{|c|}{Dataset} & \multicolumn{2}{|c|}{CIFAR-10-LT} & \multicolumn{2}{|c|}{CIFAR-100-LT} \\
    \hline
    \rowcolor{LightGray}
    \multicolumn{ 2}{|c|}{Imbalance Ratio} & 50 & 100 & 50 & 100 \\
    \hline \hline
    \multicolumn{ 2}{|c|}{\emph{\small{Resampling}}} & & & & \\
    PB Resampling & PB UBRs (\textbf{Ours}) & \textbf{0.08} & \textbf{0.28} & \textbf{0.00} & \textbf{0.09} \\
    \hline \hline
    \multicolumn{ 2}{|c|}{\emph{\small{Reweighting}}} & & & & \\
    Class-balanced Focal Loss~\cite{cui2019classbalancedloss} & UBRw Focal Loss (\textbf{Ours}) & 1.00 & 0.87 & \textbf{0.00} & \textbf{0.00} \\
    Focal Loss~\cite{focal} & UBRw Focal Loss (\textbf{Ours}) &  \textbf{0.00} & \textbf{0.00} &  \textbf{0.00} & \textbf{0.00} \\
    \hline \hline
    \multicolumn{ 2}{|c|}{\emph{\small{Margin Adjustment}}} & & & & \\
    DRO-LT & UBM DRO (\textbf{Ours}) & 0.74 & \textbf{0.22} & \textbf{0.09} & \textbf{0.12} \\
    \hline \hline
    \multicolumn{ 2}{|c|}{\emph{\small{Multi-stage Methods}}} & & & & \\
    CB Resampling~\cite{kang2019decoupling} & UBRs (\textbf{Ours}) & \textbf{0.00} & \textbf{0.00} & \textbf{0.00} & \textbf{0.00} \\
    Class-balanced Focal Loss~\cite{cui2019classbalancedloss} & UBRw Focal Loss (\textbf{Ours}) & 1.00 & \textbf{0.37} & \textbf{0.00} & \textbf{0.00} \\
    CSCE~\cite{csce} & UBRw (\textbf{Ours}) & 1.00 & 1.00 & \textbf{0.00} & \textbf{0.00} \\
    \hline
    \end{tabular}}
\end{table}
\AB{We analyze the statistical significance of our experimental results (see Table~{\RNum{1}} in the manuscript) on mitigating class imbalance with predictive uncertainty in four different setups, namely one-stage resampling, one-stage reweighting, margin adjustment and multi-stage methods. We perform the analyses according to two criteria: (i) if multiple runs are available, we compare the best method among all we reproduce, and (ii) else, the best counterpart method employing class cardinality with ours for each dataset.}

\AB{We adopt Almost Stochastic Order (ASO)~\cite{del2018optimal,dror2019deep} as implemented by~\cite{ulmer2022deep} with confidence level $0.95$. We provide $\epsilon_{min}$ values for our comparisons in~\cref{tab:sig}, which reflects the amount of violation of the stochastic order, where the lower is the better.}

\subsection{Experiments on Real-World Data}
\AB{To provide a baseline on real-world data, we performed experiments with the best mitigation methods (see Table~{\RNum{1} in the main manuscript) on the Places-LT dataset~\cite{liu2019oltr}. To be consistent with the setup presented in the manuscript, we adopted the setup of baseline training in~\cite{liu2019oltr} with the ResNet-152 architecture as the backbone and a simple classifier. \cref{fig:places-unc} compares the class cardinality- and uncertainty-based measures for Places-LT with classes.}
\begin{figure}
    \centering
    \includegraphics[width=0.4\textwidth]{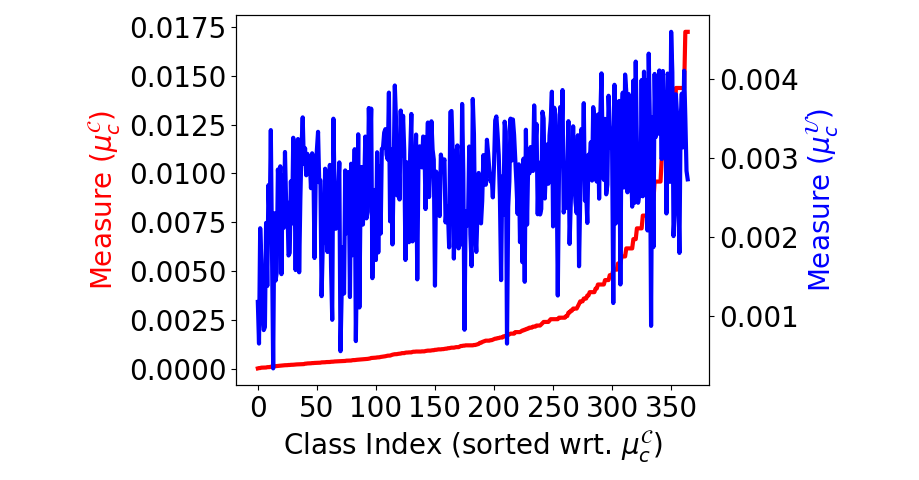}
    \caption{Comparison of class cardinality- and uncertainty-based measures for classes in Places-LT.}
    \label{fig:places-unc}
\end{figure}
\AB{Using the class uncertainties in~\cref{fig:places-unc} as our measure, we performed the experiments in~\cref{tab:places-res}. The results in the table suggest that the use of our measure significantly improves the performance of different imbalance mitigation methods. One exception is ``UBRw Focal" (uncertainty-based one-stage reweighting combined with Focal Loss), which is likely due to the two hyper-parameters of Focal Loss not being optimal in Places-LT.}
\begin{table}[htbp!]
    \caption{Top-1 error rate on Places-LT dataset. All mitigation methods improve over the na\"ive classifier.}
    \label{tab:places-res}
\centering
    \tabcolsep=2mm
    \begin{tabular}{|c|c|}
    \hline
    \rowcolor{Gray}
    Dataset & Places-LT \\
    \hline
    Na\"ive Cls. & 67.41±0.11 \\
    \hline\hline
    \emph{\small{Resampling}} & \\
    PB UBRs (\textbf{Ours}) & 60.92 ±0.09 \\
    \hline \hline
    \emph{\small{Reweighting}} & \\
    UBRw (\textbf{Ours}) & 66.54 ±0.18 \\
    UBRw Focal Loss (\textbf{Ours}) & 68.96 ±0.17 \\
    \hline \hline
    \emph{\small{Multi-stage Methods}} & \\
    UBRs (\textbf{Ours}) & \textbf{59.59 ±0.12} \\
    UBRw (\textbf{Ours}) & 66.36 ±0.07 \\
    \hline
    \end{tabular}
\end{table}

\bibliographystyle{IEEEtran} 
\bibliography{ms}